\documentclass{article}

\usepackage{tikz}
\usepackage{subcaption}%\documentclass[t,9pt]{article}
\usepackage[frenchb,english]{babel} 
\usepackage[utf8]{inputenc}
\usepackage{amsmath,amssymb,amsthm,mathrsfs,amsfonts,dsfont} 
\usepackage{graphicx}
\usepackage{xcolor}
\usepackage{mathtools}
\usepackage{algorithm,algorithmic}
\usepackage{xspace} 
\usepackage{xifthen}
\usepackage{url}
\usepackage{bbm}
\usepackage{hyperref}

\newcommand{\enm}[1]{\ensuremath{#1}\xspace}
\newcommand{\vect}[1]{{\enm{\boldsymbol{\mathbf{#1}}}}}
\renewcommand{\vec}[1]{\vect{#1}}
\newcommand{\mean}{\mathbb{E}}
\newcommand{\var}{\mathbb{VAR}}

\newcommand{\indic}{\mathbbm{1}}
\newcommand{\independ}{\perp\!\!\!\!\perp}

\newcommand{\xtarg}{\vec{x_{\text{targ}}}}
\newcommand{\EI}{\text{EI}\xspace}
\newcommand{\DOE}{\text{DoE}\xspace}
\newcommand{\EFI}{\text{EFI}\xspace}
\newcommand{\EFISUR}{\texttt{EFISUR}\xspace}
\newcommand{\EFIRand}{\texttt{EFIrand}\xspace}
\newcommand{\CEIDev}{\texttt{cEIdevNum}\xspace}
\newcommand{\Random}{\texttt{Random}\xspace}
\newcommand{\DN}{\ifmmode{\text{D\!N}}\else\text{D\!N}\xspace\fi}

\newtheorem{prop}{Proposition}

\title{A Sampling Criterion for Constrained Bayesian Optimization with Uncertainties}
\date{24 november 2023}
\author{Reda El Amri\thanks{Ecole Centrale de Lyon, now with IFP Energies Nouvelles, France}, Rodolphe Le Riche\thanks{CNRS LIMOS (Mines St-Etienne and UCA), France}, Céline Helbert\thanks{Universit\'e de Lyon - CNRS, UMR 5208, Institut Camille Jordan - Ecole Centrale de Lyon, France}, 
\\Christophette Blanchet-Scalliet\footnotemark[3], S\'ebastien Da Veiga\thanks{Safran Tech, Modelling \& Simulation, Magny-Les-Hameaux, France}
} 
\vspace{0.7cm}

\makeatletter
\usepackage{totcount}

\graphicspath{{./}}

\begin{document}
\maketitle

\begin{abstract}
We consider the problem of chance constrained optimization  
where it is sought to optimize a function and satisfy constraints, both of which are affected by uncertainties. 
The real world declinations of this problem are particularly challenging because of their inherent computational cost.\\
To tackle such problems, we propose a new Bayesian optimization method. 
It applies to the situation where the uncertainty comes from some of the inputs, 
so that it becomes possible to define an acquisition criterion in the joint 
{optimized-uncertain} input space. 
The main contribution of this work is an acquisition criterion that accounts for both the average improvement in objective function and the constraint reliability. 
The criterion is derived following the Stepwise Uncertainty Reduction logic and its maximization provides both optimal 
{design variables and uncertain} parameters.
Analytical expressions are given to efficiently calculate the criterion. Numerical studies on test functions are presented. It is found through experimental comparisons with alternative sampling criteria that the adequation between the sampling criterion and the problem contributes to the efficiency of the overall optimization. 
\end{abstract}

%%%%%%%%%%%%%%%%%%%%%%%%%%%%%%%%%%%%%%%%%%%%%%%%%%%%%%%%
%%%%%%%%%%%%%%%%%%%%%%%%%%%%%%%%%%%%%%%%%%%%%%%%%%%%%%%%
\vskip\baselineskip
\noindent Please cite as : to appear in SMAI Journal of Computational Mathematics, Vol.9, pp. 285-309, https://doi.org/10.5802/smai-jcm.102, 2023.
\vskip\baselineskip
\href{https://creativecommons.org/licenses/by/4.0/}{\includegraphics[height=0.5cm]{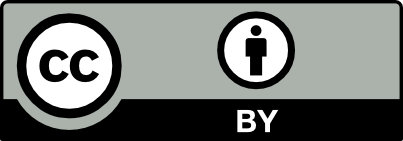}}~This work is licensed under a Creative Commons Attribution 4.0 International License.
%%%%%%%%%%%%%%%%%%%%%%%%%%%%%%%%%%%%%%%%%%%%%%%%%%%%%%%%
\newpage
\section{Introduction}
%\rodo{il y a maintenant la commande \texttt{\textbackslash{rev\{\}}} à utiliser pour signaler les changements dans le papier après révision. Ainsi, les changements validés passent de \texttt{\textbackslash{nameadd\{\ldots\}}} à \texttt{\textbackslash{rev\{\ldots\}}} }\\
Despite the long-term research efforts put into numerical optimization, many practical applications remain difficult. There are three main reasons: most real problems involve nonlinear constraints, the objective function or the constraints are numerically costly to evaluate (e.g., when nonlinear finite elements underlie the optimization criteria), and some of the parameters are uncertain.

To ease the computing load, Bayesian Optimization (BO) incorporates kriging surrogates to save calls to the objective function, as embodied in the archetypal EGO algorithm \cite{schonlau1998global}. 
The original optimization problem is translated into a series of other problems, that of the acquisition of new points where the costly function will be calculated. The acquisition criterion is based on the kriging model and it mitigates the optimization of the function and the improvement of the kriging model {\cite{garnett2022}}. 
BO has rapidly been extended to encompass constraints \cite{sasena2002global,EFIPicheny}{\cite{Erikson2021,hernandez2015}}.

\vskip\baselineskip
In this article, the focus is not only on costly and general nonlinear constrained optimization but also on problems that are affected by uncertainties. 

Uncertainties may originate from random environments such {as weather}, noise in sensors or uncontrolled boundary conditions. 
Many physical models also come with uncertainties in an attempt to describe a lack of knowledge about the true phenomena. 
Big data applications are confronted to uncertainties which reflect the part of the data that cannot be handled at a time, either 
because the data arrive as a dynamic flow, or because the volume is too large to be processed in a single step.
Since uncertainties are so ubiquitous, robustness against uncertainties is becoming an important aspect of optimization problem solutions.

When the uncertainties cannot be characterized in stochastic terms such as probabilities, 
strong guarantees about the robustness of the solutions can be obtained with deterministic approaches based on worst-case {scenarios} over the set of possible uncertainties \cite{ben2009robust,gabrel2014recent}{\cite{marzat2016}}.  
If this set is large (possibly infinite) and the problem non-convex, the conservatism of the solutions and the computational tractability are inherent difficulties of this family of methods.
When the uncertainties are seen as stochastic, two situations may be distinguished. 
%They are introduced below in the context of the optimization of a system designed to work in an environment. 
%\chris{je ne suis pas sure de bien comprendre cette dernière phrase. Suggestion: They are introduced below in the context of the optimization.}\rodo{juste en dessous on utilise les mots ``manufacturing'', ``environment'', ``system'', qui ne sont pas généraux mais liés au cas d'un problème de design d'un objet/système qui fonctionnera dans un environnement ensuite, c'est pour introduire cela.} 
% \rodo{finalement je crois qu'on peut se passer de cet exemple}

{In a first} class of problems, the uncertainties are {instantiated} within the objective or constraint functions and cannot be chosen. 
Such uncertainties are {a hard coded, not controllable,} noise corrupting the objective and the constraint functions.
A typical situation is when the functions resorts to random number generations, e.g., for a Monte Carlo simulation, and no access to the source code is granted.
%\rodo{Pour continuer notre discussion: je crois qu'en fait cette catégorie est quand on ne peut pas choisir les $u$ dans la simulation. Ils sont subis à travers $f(x,u)$, d'où le rapprochement avec les codes stochastiques. Si l'idée était que $u$ et $x$ se mélangent pour définir des variables effectives, comme pour les tolérances de fabrication, je crois qu'on se ramène à l'autre catégorie en version simplifiée: par exemple $x$ défini les paramètres de la loi de $u$ et la fonction est $f(u|x)$: en mécanique, $f$ = propriété d'une structure définie par sa dimension, $x$ = dimension théorique et classe de tolérance, $u$ = dimension effective. Mais on peut choisir $u$ dans les calculs -- pas en vraie fabrication --, donc on peut évaluer $f(u)$, donc on peut kriger dans l'espace des $(x,u)$ qui se replie sur l'espace des $u$ seulement. Ma proposition: 1ère classe = pb stochastiques où $u$ n'est pas choisi, 2nde classe = pb où $u$ peut être choisi dans les simulations. Si vous êtes ok il faudra le dire plus clairement}
Stochastic algorithms can, under conditions about their own stochasticity, accomodate the noise in the observations and still converge to problem solutions: 
this has given rise in the early 50's to stochastic descent algorithms \cite{kiefer1952stochastic,arrow1958studies} 
that have since then experienced great developments \cite{spall2005introduction,andrieu2011gradient} often in relation 
to machine learning \cite{kingma2014adam}; 
well-performing versions of (stochastic) evolutionary algorithms for noisy functions have been identified \cite{brockhoff_miror_2010,loshchilovsaACM2012} thanks to competitions on reference benchmarks \cite{COCO2012noisy}.

{In a second} class of problems, the uncertainties perturb parameters that are distinct from the optimization variables and can be chosen during the simulations. 
The separation between {optimization (or design)} variables and uncertain parameters was already underlying Taguchi's design of experiments in the 80's \cite{logothetis1989quality}.
Because of this separation and providing that a probability of occurence of the uncertainties exists, a statistical modelling in the joint \emph{{design} $\times$ uncertain} parameters space is possible. 
This will be the context of the current work.

A key step when optimizing in the presence of uncertainties is the formulation of the problem, i.e., the choice of the robustness criteria.
Considering first unconstrained problems, relevant criteria are the expectation of the objective \cite{janusevskis_jogo_2012} or one of its (super-)quantiles \cite{torossian2019mathcal,torossian2020review}.
%\rododelete{or the worst-case}\rodo{mis plus haut dans les approches déterministes et pas forcément défini}
In Robust Optimization, the uncertainties are handled in terms of specific compromises between the average performance and its dispersion \cite{park_robust_AIAA_2006,beyer2007robust} or by investigating all such Pareto optimal compromises through a multi-objective formulation \cite{RIBAUD2020106913}.

When there are constraints that depend on the uncertainties, the feasibility of the solutions is typically measured in probability. 
Probabilistic models of the constraints are called chance constraints \cite{nemirovski2012safe} or reliability constraints \cite{bourinet2018reliability}. %\cite{fauriat2014ak}.
The field of Reliability-Based Design Optimization (RBDO) is concerned with the resolution of optimization problems that contain reliability constraints \cite{balesdent2020overview}. 
The optimization problems are formulated in terms of statistical criteria such as probabilities of satisfying the constraints, expectations or (super-)quantiles or conditional quantiles of the objective function \cite{torossian2019mathcal,leriche:cel-02285533,pujol2009incertitude}.
When the statistical criteria cannot be calculated analytically, the bottleneck of the computational cost of the optimization is even more stringent since the statistical criteria must be numerically estimated within the optimization iterations.
This is sometimes called the double loop issue { i.e., for each design point visited, many simulations are needed to encompass the effects of the uncertainties}.
Several paths have been proposed to circumvent the double loop issue, some approximating the probability of feasibility by reliability indices \cite{valdebenito2010survey}, others improving the efficiency of the probability calculations (e.g., stratified sampling in \cite{zuniga2012analysis}), others decoupling the reliability estimation from the optimization. \cite{schueller2008computational} gives a review of some of these techniques. 

In the last decade, numerous contributions to the optimization of costly functions with uncertainties have relied on the learning of a metamodel of the true functions, in particular Gaussian processes (GP) \cite{dubourg2011reliability,moustapha2016quantile}. 
In \cite{janusevskis_jogo_2012} and \cite{amri:hal-02986558}, the GP not only helps for the optimization (or for the inversion) of the design variables, but it also serves to define an optimal sampling scheme.

In this article, the problem of minimizing the mean of a stochastic function under 
chance constraints is addressed. The objective function and the constraints are costly in the sense that they cannot be calculated more than a hundred times. 
Furthermore the problem is assumed to be nonconvex so that part of the solution process will not 
be analytical. 
Uncertainties are described by parameters different from the optimization variables and that can be chosen in the calculations. 
Generalizing \cite{janusevskis_jogo_2012}, an optimization and sampling Bayesian procedure is proposed that accounts for probabilistic constraints.
{
The acquisition criterion is based on the feasible improvement, which is known to be more myopic (less explorative) but also numerically more tractable than its information theoretic counterparts \cite{hernandez2015,perrone2019constrained}. Indeed, information theoretic criteria, which quantify reductions in the entropy of the predicted optima locations or values, imply internal optimizations of the GP trajectories. Note also that \cite{hernandez2015,perrone2019constrained} do not consider uncertain parameters. 
A less myopic feasible improvement criterion, which looks two-steps ahead, is given in \cite{zhang2021constrained}, yet, again, this work does not address uncertain parameters.
On the contrary, uncertainties are examined in \cite{qing2022spectral}: thanks to a spectral decomposition of the GP covariance functions, 
 the variance and the mean of the objective function under the uncertain parameters become a GP, which then opens the way to  acquisition criteria based on improvement. The constraint bears on the objective variance. Our contribution differs from \cite{qing2022spectral} essentially by the proposition of a policy to choose where to evaluate the uncertain parameters, and secondarily, by the consideration of general constraints.}

After formulating the problem (Sections \ref{sec-PbFormulation} and \ref{sec-GP}), a principle for devising robust bayesian problems is stated (Section \ref{sec-generalBOprc}) which applies to any given progress measure. 
In Section \ref{sec-xtarg}, this principle is applied to the feasible improvement as a specific progress measure. The associated sampling criterion is introduced in Section \ref{SamplingCrit}. It is a Stepwise Uncertainty Reduction (SUR) criterion \cite{bect2012sequential}, a one-step-ahead variance reduction, for which a proxy which is easier to compute is presented. The resulting algorithm is summarized in Section \ref{sec:numerical}, and its performance assessed on an analytical and an industrial test case. 
%\chrisdelete{For completeness, a derivation of the variance of the improvement (different from the derivation given in \cite{schonlau1998global})   can be found in Appendix.} \rodo{on est d'accord que ce n'est pas la même preuve?}\chris{non c'est essentiellement la même preuve. Personnellement du coup je ne le mentionnerai pas dans l'intro}

%%%%%%%%%%%%%%%%%%%%%%%%%%%%%%%%%%%%%%%%%%%%%%%%%%%%%%%%
 
\section{Problem formulation}

%%%%%%%%%%%%%%%%%%%%%%%%%%%%%%%%%%%%%%%%%%%%%%%%%%%%%%%%
%%%%%%%%%%%%%%%%%%%%%%%%%%%%%%%%%%%%%%%%%%%%%%%%%%%%%%%%

\subsection{Starting problem}
\label{sec-PbFormulation}

Let $f(\vec{x},\vec{u})$ be the scalar output of an expensive computer simulation and let $g_i(\vec{x},\vec{u})$, $i=1,\dots,l$ be the set of  constraints,  where $\vec{x} \in \mathcal{S_X}$ can be precisely chosen while $\vec{u} \in \mathcal{S_U}$ is a realization of a vector of random variables $\vec{U}$ with a specified probability density function $\rho_\vec{U}$. Such a formulation with design and random variables is general to all optimization problems under uncertainty. 
%It was for example already underlying Taguchi's design of experiments in the 80's \cite{logothetis1989quality}.\rodo{déplacé en intro}
%\\\rodo{je change un peu le texte pour généraliser notre formulation}\\
The examples in this article belong to continuous optimization in the sense that $\mathcal{S_X} \subset \mathbb{R}^d$ and $\mathcal{S_U} \subset  \mathbb{R}^m$. 
Nevertheless, it is important to note that the framework of our work, the Gaussian processes and the algorithms which will be introduced, generalize nicely to spaces that contain discrete variables (see for example \cite{pelamatti2019efficient,roustant2019}).

Our goal is to find $x$ which minimizes $f$ while insuring that the $g_i$'s lies under a failure threshold (0 in general). 
In the presence of uncertainties, Robust Optimization aims at controlling the impact of uncertainties on the performance of the optimal solution. 
%\rododelete{In a probabilistic framework for uncertainty analysis,}\rodo{déjà dit plus haut avec la densité $\rho_\vec{U}$} 
The problem is that $f(\vec{x},\vec{U})$ and $g_i(\vec{x},\vec{U})$, $i=1,\dots,l$  are  random quantities induced by $\vec{U}$. 
In order to perform optimization, we need to fall back on a deterministic form which is achieved by applying some statistical measures to $f(\vec{x},\vec{U})$ and $g_i(\vec{x},\vec{U})$, $i=1,\dots,l$ \cite{torossian2019mathcal,leriche:cel-02285533}. 

In this article, the constrained optimization problem under uncertainties is formulated as the minimization of the expectation over \vec{U} of $f$ while all the constraints $g_i(\vec{x},\vec{U}) \leq 0$, $i=1,\dots,l$ are satisfied with a high probability:
\begin{equation}
\begin{split}
\vec{x}^*  = &\arg \min\limits_{\vec{x} \in \mathcal{S_X}} ~\mean[f(\vec{x},\vec{U})] ~\text{s.t.}~ \mathbb{P}(g_i(\vec{x},\vec{U}) \leq 0,~ i=1,\dots,l ) \geq 1- \alpha \\
& \text{where } \vec{U} \sim \rho_{\vec{U}} \text{ with support } \mathcal{S_U} . 
\end{split}
\label{problem}
\end{equation}
$\alpha$ is a reliability parameter representing the allowed constraint violation level ($0 < \alpha < 1$). 

In the formulation of Equation~(\ref{problem}), the emphasis is on constraint satisfaction with a guaranteed reliability. 
This is common in RBDO where constraints are satisfied in probability and the objective 
function is deterministic \cite{dubourg2011reliability}. Such problems are also said to have chance constraints \cite{ben2009robust}. In addition in Equation~(\ref{problem}), random events affect the objective function and 
are taken into account through an expectation. Thanks to its linearity, the expectation is the simplest statistical measure. 
Besides, efficient approaches exist to estimate and optimize it \cite{janusevskis_jogo_2012}. 
Formulations such as Equation~(\ref{problem}) with a mean objective function and chance constraints have been called a ``model for qualitative failure'' \cite{andrieu2011gradient} in which the objective function and the constraints are taken as independent. 
%\celidelete{Another natural statistical measure for stochastic objective functions is the quantile \cite{torossian2020review,torossian2019mathcal} because the optimization result, whether an objective function or a constraint, has an associated confidence level.}\rodo{pourquoi enlever cette phrase?}\rodo{biblio mise en intro}
Other formulations with quantiles conditioned by feasibility have been given in \cite{leriche:cel-02285533,pujol2009incertitude} but they remain an open challenge for costly problems. In the current work, Equation~(\ref{problem}) is addressed because it is a compromise  between complete robust formulations and mathematical tractability. 
%\rodo{ Il doit aussi y avoir d'autres papier en finance avec des CVAR/super-quantiles/expectiles et ce genre de choses qu'on pourrait citer, si vous avez une idée...}\chris{j'ai fait un peu de recherche et je n'ai pas trouvé des choses qui étaient vraiment en lien avec notre formulation car la CVar c'est $E[x|X<F_X^{-1}(\alpha)]$. Du coup je trouev que c'ets difficile de faire le lien}

By seeing the probability as an expectation, the constraint part of Equation \eqref{problem} becomes
\begin{equation*}
\begin{split}
\mathbb{P}(g_i(\vec{x},\vec{U}) &\leq 0,~ i=1,\dots,l ) \geq 1 - \alpha \\ 
%& \Leftrightarrow \mean_\vec{U}[ \indic_{\{g_i(\vec{x},\vec{U}) \leq 0,~ i=1,\dots,l \}}]  \geq 1 - \alpha\\
 & \Leftrightarrow 1 - \alpha -\mean[\indic_{\{g_i(\vec{x},\vec{U}) \leq 0,~ i=1,\dots,l \}}]  \leq 0.
\end{split}
\end{equation*}
%\rodo{j'ai compactifié l'écriture ci-dessus et utilisé la notation $\coloneqq$ pour désigner une définition}\\
From the last expression, the problem (\ref{problem}) is equivalent to 
\begin{equation}
\vec{x}^* = \arg \min\limits_{\vec{x} \in \mathcal{S_X}} ~z(\vec{x}) ~\text{s.t.}~ c(\vec{x}) \leq 0
\label{newproblem}
\end{equation}
where $z(.) \coloneqq \mean[f(.,\vec{U})]$ and $c(\vec{x}) \coloneqq 1 - \alpha -\mean[\indic_{\{g_i(\vec{x},\vec{U}) \leq 0,~ i=1,\dots,l \}}]$.

%%%%%%%%%%%%%%%%%%%%%%%%%%%%%%%%%%%%%%%%%%%%%%%%%%%%%%%%
%%%%%%%%%%%%%%%%%%%%%%%%%%%%%%%%%%%%%%%%%%%%%%%%%%%%%%%%

\subsection{Gaussian Process regression framework}
\label{sec-GP}
In the context of expensive computer simulation, the Problem (\ref{newproblem}) is approximated with Gaussian processes (GPs).  
Directly building a metamodel for $z$ and $c$ would need too many evaluations of $f$ and $g_i$ to estimate the expectation and the probabilities. 
Therefore, GP approximations of $f$ and the $g_i$'s are built in the joint space $\mathcal{S_X} \times \mathcal{S_U}$. 
Models for $z$ and $c$ in the design space $\mathcal {S_X}$ are then deduced from them. 
More precisely, we suppose that $f$ and the constraints $(g_i)_{i=1}^{l}$ are realizations of independent Gaussian processes $F$ and $G_i$ such that
\begin{equation*}
\begin{split}
F{(\vec{x},\vec{u})} &\sim \mathcal{GP}(m_F(\vec{x},\vec{u}) , k_F(\vec{x},\vec{u},\vec{x'},\vec{u'}) ),\\
\forall i = \{1,\dots,l\} ~,~ G_i{(\vec{x},\vec{u})} &\sim \mathcal{GP}(m_{G_i}(\vec{x},\vec{u}) , k_{G_i}(\vec{x},\vec{u},\vec{x'},\vec{u'}) ),
\end{split}
\end{equation*}
where $m_F$ and $m_{G_i}$ are the mean functions while $k_F$ and $k_{G_i}$ are the covariance functions. \\
Let $F^{(t)}$ and $G_i^{(t)}$ denote the Gaussian processes conditioned on the t observations, $f^{(t)} = (f(\vec{x}_1,\vec{u}_1),\ldots, f(\vec{x}_t,\vec{u}_t))$ and  $g_i^{(t)} = (g_i(\vec{x}_1,\vec{u}_1),\ldots, g_i(\vec{x}_t,\vec{u}_t))$, obtained at points 
$D^{(t)} = \{(\vec{x}_k,\vec{u}_k)~,~k=1,..,t\}$. Since the expectation is a linear operator applied to $f$, it follows that {$Z^{(t)}(\vec{x}) =\int_{\mathbb{R}^m}F^{(t)}{(\vec{x},\vec{u})}\rho_\vec{U}(\vec{u}) d\vec{u}$} is still a Gaussian process with known mean $m^{(t)}_Z$  and covariance function $k^{(t)}_Z$ given by:

\begin{equation}
\begin{split}
m^{(t)}_Z(\vec{x}) &= \int_{\mathbb{R}^m} m^{(t)}_F(\vec{x},\vec{u}) \rho_\vec{U}(\vec{u}) d\vec{u}, \\
k^{(t)}_Z(\vec{x},\vec{x'}) &= \iint \limits_{\mathbb{R}^{2m}} k^{(t)}_F(\vec{x},\vec{u},\vec{x'},\vec{u'}) \rho_\vec{U}(\vec{u},\vec{u'}) d\vec{u} d\vec{u'}.  \\
\end{split}
\label{ZGP}
\end{equation}
The integrals that appears in (\ref{ZGP}) can be evaluated analytically for specific choices of $\rho_U$ and $k_F$ (see \cite{janusevskis_jogo_2012}). In the general case, a quadrature rule can be used to approximate these integrals {or a Monte Carlo method if $m$ is large}.   

We also introduce the process 
{
\begin{equation*}
C^{(t)}\!(\vec{x}) = 1 - \alpha -\int_{\mathbb{R}^m} \indic_{{\cap_{i=1}^l}\{G_i^{(t)}\!(\vec{x},\vec{u}) \leq 0 \}}\rho_\vec{U}(\vec{u}) d\vec{u}
\label{eq:Cofx}
\end{equation*}
}
which is the statistical model of the constraint $c$ (Equation (\ref{newproblem})). 
Note that the process $C^{(t)}$ is not Gaussian.

{To solve Problem (\ref{newproblem}), Bayesian optimization is used where functions $z$ and $c$ are considered realizations of processes $Z$ and $C$. The procedure is detailed next.}

%%%%%%%%%%%%%%%%%%%%%%%%%%%%%%%%%%%%%%%%%%%%%%%%%%%%%%%%
%%%%%%%%%%%%%%%%%%%%%%%%%%%%%%%%%%%%%%%%%%%%%%%%%%%%%%%%

\subsection{A general principle for devising robust Bayesian optimization algorithms}
\label{sec-generalBOprc}
Now that the problem has been formulated (Section~\ref{sec-PbFormulation}) and the GP models introduced (Section~\ref{sec-GP}), we present a Bayesian algorithm to solve Problem~(\ref{problem}) within a restricted number of calls to $f$ and the $g_i$'s. 
Before going into the details of the method, it is important to understand the general principle that underlies the design of robust Bayesian Optimisation (BO) algorithms.

\begin{prop}[A general principle to devise robust BO algorithms]
\label{prop:general_prc}
~\vskip 0.1cm
Robust Bayesian Optimization algorithms can be designed as follows:\\
\emph{A)} define a progress measure $P(x)$ in relation with the problem formulation and calculated from the GPs trajectories. \\
\emph{B)} The robust BO algorithm is:
\begin{algorithmic}
\STATE Define an initial space filling design in the joint space $\mathcal{S_X}\times\mathcal{S_U}$ : $D^{(n_0)}$ 
\STATE Initialize all conditional GPs: $F^{(n_0)}$, $G^{(n_0)}_i,i=1,\ldots,l$ with $D^{(n_0)}$
\WHILE{stopping criterion not met}
\STATE Determine a desirable, targeted, \vec{x} by maximizing the expectation of the progress measure, 
\begin{equation}\label{eq:progress}
\xtarg = \arg \max_{\vec{x} \in \mathcal{S_X}} \mean \left( P^{(t)}(\vec{x}) \right)~.
\end{equation}
\STATE The next iterate minimizes the one-step-ahead variance of the progress measure at $\xtarg$,
\begin{equation}\label{eq:varprogress}
(\vec{x}_{t+1},\vec{u}_{t+1}) ~=~ \arg \min_{\tilde{\vec{x}},\vec{\tilde{u}} \in \mathcal{S_X} \times \mathcal{S_U}} \mathbb{VAR} \left( P^{(t+1)}(\xtarg) \right)~
\end{equation}
where $P^{(t+1)}$ is evaluated with GPs updated according to $D^{(t+1)}= D^{(t)} \cup \{(\tilde{\vec{x}},\vec{\tilde{u}})\}$.
\STATE Calculate the simulator response, i.e., $f$ and $g_i,i=1,\ldots,l$, at the next point ($\vec{x}_{t+1},\vec{u}_{t+1}$).
\STATE Update the design $D^{(t+1)}= D^{(t)} \cup \{(\vec{x}_{t+1},\vec{u}_{t+1})\}$
\STATE Update the Gaussian processes, $F^{(t+1)}$, $G^{(t+1)}_i,i=1,\ldots,l$.
\ENDWHILE
\end{algorithmic}
\end{prop}

Various algorithms can be obtained by changing the measure of progress $P$.
In this article it will be the feasible improvement, which will soon be presented.
Other measures of progress are possible, for example
$P(\vec{x}) = -F(\vec{x}) - \sum_{i=1}^{l}p_i\max(0~,~G_i(\vec{x},\vec{u}^{\text{mod}}))$ 
where the $p_i$'s are positive penalty parameters and $ \vec{u}^{\text{mod}}$ the mode of \vec{U}, or $P(\vec{x}) = \max\left([z_\text{min}^\text{feas}-Z(\vec{x}) \mid C(\vec{x}) \le 0] , 0\right)$ where $z_\text{min}^\text{feas}$ is the best objective function associated to a feasible point. 
The goal of the next sections is to present the methodology and the associated formulas when the progress measure is the feasible improvement, 
chosen for its closeness to the problem formulation and its computability.
The one-step-ahead variance can be difficult to tackle so approximations are useful.  
In this text, the generic term ``sampling criterion'' relates to the one-step-ahead variance or its proxy. 
For costly problems, the stopping criterion in Proposition~\ref{prop:general_prc} is often a given number of calls to the simulator. 

%\rodo{expl d'alternative ci-dessous}\chris{pour quoi l'appeler $I_c$ et pas $P$. Cela me prait un peu bizarre de donner des expression pour d'autres $P$ alors qu'on en donne pas l'expression de celle que l'on va utiliser. Perso la phrase du début de la section me convient. J'ai preu qu'on perde les lecteurs avec ces autres formulations}\rodo{ok pour $P$, j'ai changé. Si on dit que c'est un principe général il faut donner d'autres cas que celui du papier d'où les 2 lignes ci dessous\chris{OK mais alors je les mets au dessous juste après la première phrase. Je trouve qu'on perd le lecteur à revenir après notre choix}}
%\chrisdelete{Other robust BO algorithms could be obtained with progress measures such as $P(\vec{x}) = -F(\vec{x}) - \sum_{i=1}^{l}p_i\max(0~,~G_i(\vec{x},\vec{u}^{\text{mod}}))$ where the $p_i$'s are positive penalty parameters and $ \vec{u}^{\text{mod}}$ the mode of \vec{U}, or $P(\vec{x}) = \max\left([z_\text{min}^\text{feas}-Z(\vec{x}) \mid C(\vec{x}) \le 0] , 0\right)$ where $z_\text{min}^\text{feas}$ is the best objective function associated to a feasible point. 
%A lot of options are possible for $P$ but the current article will only discuss one of them, chosen for its closeness to the problem formulation and its computability.}

%%%%%%%%%%%%%%%%%%%%%%%%%%%%%%%%%%%%%%%%%%%%%%%%%%%%%%%%
%%%%%%%%%%%%%%%%%%%%%%%%%%%%%%%%%%%%%%%%%%%%%%%%%%%%%%%%

\section{The progress measure and the associated targeted point $\xtarg$}
\label{sec-xtarg}
Following Proposition \ref{prop:general_prc}, the first step consists in defining a progress measure $P^{(t)}$, which will be the cornerstone of the definition of what a most promising candidate for evaluation, $\xtarg$, is. 
The maximization of its expectation should contribute to both solving the constrained optimization problem and improving the GPs.
The most popular progress measure for optimization under constraints is the Feasible Improvement \cite{schonlau1998global,sasena2002exploration} defined by   
\begin{equation*}
FI^{(t)}(\vec{x}) = I^{(t)}(\vec{x}) ~ \indic_{\{C^{(t)}(\vec{x}) \leq 0\}}~,
\end{equation*}
where $I^{(t)}(x)=\big(z_{\min}^{\text{feas}} - Z^{(t)}(\vec{x})\big)^+$ denotes the improvement over the current feasible minimum value.

In our case, $z_{\min}^{\text{feas}}$ must be further explained because it is not directly observed. 
This will be the subject of the next section.
The definition of $z_{\min}^{\text{feas}}$ and the fact that $C^{(t)}(\vec{x})$ is not Gaussian is a difference between the $FI$ of this article and those in \cite{schonlau1998global,sasena2002exploration}.

Following Proposition \ref{prop:general_prc} and Equation \eqref{eq:progress} the promising point in the control space is obtained by maximizing the expectation of the progress measure. Here it corresponds to maximizing the \emph{Expected Feasible Improvement} (\EFI),
\begin{equation}
\xtarg = \arg \max \limits_{\vec{x} \in \mathcal{S_X}} {\EFI}^{(t)}(\vec{x}).
\label{problemEFI}
\end{equation}
where ${\EFI}^{(t)}(\vec{x})$ is $\mean \left( FI^{(t)}(\vec{x}) \right)$.
The independence of the GP processes implies that the \EFI can be  expressed as
 \begin{equation*}
 \begin{split}
\EFI(\vec{x}) &= \EI^{(t)}(\vec{x}) \mathbb{P}(C^{(t)}(\vec{x}) \leq 0).
\end{split}
\label{EFI}
\end{equation*}
The first term is the well known \emph{Expected Improvement} (\EI) for which an analytical expression is available,
\begin{equation}\label{eq:EI}
\EI^{(t)}(\vec{x}) = (z_{\min}^\text{feas} - m_Z^{(t)}(\vec{x})) \Phi\bigg(\frac{z_{\min}^\text{feas} - m_Z^{(t)}(\vec{x})}{\sigma_Z^{(t)}(\vec{x})}\bigg) + \sigma_Z^{(t)}(\vec{x}) \phi\bigg(\frac{z_{\min}^\text{feas} - m_Z^{(t)}(\vec{x})}{\sigma_Z^{(t)}(\vec{x})}\bigg),
 \end{equation}
where $\sigma_Z^{(t)}(\vec{x}) = \sqrt{ k^{(t)}_Z(\vec{x},\vec{x})}$, $\Phi$ and $\phi$ are the normal cumulative distribution and density functions, respectively. 
The second term, $\mathbb{P}(C^{(t)}(\vec{x}) \leq 0)$, can be approximated with available numerical methods (see details in Section \ref{sec:implemDetails}).

%%%%%%%%%%%%%%%%%%%%%%%%%%%%%%%%%%%%%%%%%%%%%%%%%%%%%%%%
%%%%%%%%%%%%%%%%%%%%%%%%%%%%%%%%%%%%%%%%%%%%%%%%%%%%%%%%

\subsection*{Definition of the current feasible minimum $z_{\min}^{\text{feas}}$}

To solve problem \eqref{problemEFI}, we need to define $z_{\min}^{\text{feas}}$ . We extend the definition of the current minimum for a non-observed process $Z$ introduced in \cite{janusevskis_jogo_2012} to a problem with constraints. $z_{\min}^{\text{feas}}$ is defined as the minimum of the mean of the process $Z^{(t)}$ such that the constraint is satisfied in expectation 
\begin{equation}\label{eq:zminfeas}
z_{\min}^{\text{feas}} =  \min_{\vec{x} \in \mathcal{X}_t} m^{(t)}_Z(\vec{x}) ~\text{s.t.}~ \mean[ C^{(t)}(\vec{x}) ] \leq 0~.
\end{equation}

Under Fubini's condition and as the constraints are conditionally independent given \vec{x} and \vec{u}, the expectation of $C^{(t)}$ is an integral over the uncertain space of a product  of univariate Gaussian cumulative distribution functions
{
\begin{equation}
\begin{split}
\mean[ C^{(t)}(\vec{x}) ] &= \mean \left[ 1 - \alpha -\int_{\mathbb{R}^m} \indic_{{\cap_{i=1}^l}\{G_i^{(t)}(\vec{x},\vec{u}) \leq 0 \}}\rho_\vec{u}(\vec{u}) d\vec{u} \right] \\
%&= 1 - \alpha - \mean_\vec{U}[\mean[ \indic_{{\cap_{i=1}^l}\{G_i^{(t)}(\vec{x},\vec{U}) \leq 0 \}}] ] \\
&= 1 - \alpha - \int_{\mathbb{R}^m}  \prod \limits_{i=1}^{l}\mean[ \indic_{\{G_i^{(t)}(\vec{x},\vec{u}) \leq 0\}} ] \rho_\vec{u}(\vec{u}) d\vec{u}\\
%&= 1 - \alpha - \int_{\mathbb{R}^m}\prod \limits_{i=1}^{l} \mathbb{P}( G_i^{(t)}(\vec{x},\vec{u}) \leq 0)\rho_\vec{U}(\vec{u}) d\vec{u}\\
&= 1 - \alpha - \int_{\mathbb{R}^m}\prod \limits_{i=1}^{l} \Phi\left(\frac{-m_{G_i}^{(t)}(\vec{x},\vec{u})}{\sigma_{G_i}^{(t)}(\vec{x},\vec{u})}\right)\rho_\vec{U}(\vec{u}) d\vec{u}.
\end{split}
\end{equation} 
}
If there is no solution to problem (\ref{eq:zminfeas}), we choose the most feasible point in expectation,
\newcommand{\xmf}{\mathbf{x}^{\text{mf}}}
\begin{equation*}
z_{\min}^\text{feas} = m_Z^{(t)}(\xmf) \text{ where } \xmf = \arg\max \limits_{\mathbf{x} \in \mathcal{X}_t}  \int_{\mathbb{R}^m}\prod \limits_{i=1}^{l} \mathbb{P}( G_i^{(t)}(\vec{x},\vec{u}) \leq 0)\rho_\vec{U}(\vec{u}) d\vec{u}.
\end{equation*}

%%%%%%%%%%%%%%%%%%%%%%%%%%%%%%%%%%%%%%%%%%%%%%%%%%%%%%%%
%%%%%%%%%%%%%%%%%%%%%%%%%%%%%%%%%%%%%%%%%%%%%%%%%%%%%%%%

\section{Extending the acquisition criterion} 
\label{SamplingCrit}
The sequential methodology introduced in Proposition \ref{prop:general_prc}, Equation \eqref{eq:varprogress}, 
requires to choose a couple $(\vec{x}_{t+1},\vec{u}_{t+1})$ such that the variance of the one-step-ahead feasible improvement is minimal, i.e.,
\begin{equation}
(\vec{x}_{t+1},\vec{u}_{t+1})= \arg \min_{(\vec{\tilde{x},\tilde{\vec{u}}}) \in \mathcal{S_X}\times \mathcal{S_U}} \var \left(I^{(t+1)}(\xtarg) ~ \indic_{\{C^{(t+1)}(\xtarg) \leq 0\}}\right),
\label{eq-VFItp1}
\end{equation}
where $I^{(t+1)}$ and $C^{(t+1)}$ are the updated $I^{(t)}$ and $C^{(t)}$ taking into account the observations at the point $(\vec{\tilde{x},\tilde{\vec{u}}})$.
This choice increases the most the information provided at the current point of interest, $\xtarg$, where the information now includes both the probabilistic constraints and the improvement.

In \cite{janusevskis_jogo_2012}, the authors noted that $\vec{x}_{t+1}$ is usually very close to $\xtarg$ so instead of looking for the couple $(\vec{x}_{t+1},\vec{u}_{t+1})$, we assume that $\vec{x}_{t+1}=\xtarg$. 
This simplifies the optimization of the sampling criterion because it reduces the dimension of the minimization (Equation~(\ref{eq-VFItp1})).
As the one-step-ahead variance of the feasible improvement is difficult to evaluate, {a computationally more tractable quantity}, built with a product of variances, is now proposed. 

%%%%%%%%%%%%%%%%%%%%%%%%%%%%%%%%%%%%%%%%%%%%%%%%%%%%%%%%
%%%%%%%%%%%%%%%%%%%%%%%%%%%%%%%%%%%%%%%%%%%%%%%%%%%%%%%%

\subsection*{The sampling criterion}
We introduce a new sampling criterion specific to the robust optimization Problem~\eqref{problem}, denoted $S$ for Sampling as it is used to generate a new value for \vec{u}. It is  {inspired by }the variance of the one-step-ahead feasible improvement {but easier to assess}. 
\begin{prop}
%[\celidelete{\chrisdelete{Aggregated variance}\chrisadd{Proxy \rodoadd{to the variance} of the feasible \rododelete{improvement variance} \rododelete{(PVFI)}\rodo{on ne se sert pas de PVFI, il y a $S$}}}]
\label{prop:samplingcrit}
 {The sampling criterion}, given a new observation at point $(\tilde{\vect{x}},\tilde{\vec{u}})$, is
\begin{equation}
\begin{split}
S&(\tilde{\vec{x}},\vec{\tilde{u}}) = \var\big( I^{t+1}(\xtarg)\big) \int_{\mathbb{R}^m} \var\big( \indic_{{\cap_{i=1}^l}\{G_i^{(t+1)}(\xtarg,\vec{u}) \leq 0 \}} \big) \rho_\vec{U}(\vec{u}) d\vec{u},\\
=&~\var(\big(z_{\min}^{\text{feas}} - Z^{(t+1)}(\xtarg)\big)^+)\int_{\mathbb{R}^m} \var\big( \indic_{{\cap_{i=1}^l}\{G_i^{(t+1)}(\xtarg,\vec{u}) \leq 0 \}} \big) \rho_\vec{U}(\vec{u}) d\vec{u}.
\end{split}
\label{SC}
\end{equation}
\end{prop}

In Equation \eqref{SC} the first part of the expression forces the sampling 
to reduce the uncertainty in the improvement value while the second part focuses on the averaged predicted feasibility variance. 
For the sake of calculation simplicity, it might seem preferable to replace the improvement by the objective process $Z^{(t+1)}$ and the variance of feasibility by the variance of the constraints. 
However, such variances are not those of the quantities of interest. In optimization, it is not important to reduce the variance of the constraints if it is clear that points are feasible or infeasible. However it is important to reduce the variance when there is a large uncertainty on feasibility, which is achieved by considering the variance of the Bernoulli variable $\indic_{{\cap_{i=1}^l}\{G_i^{(t+1)}(\xtarg,\vec{u}) \leq 0\}}$. In the same way, the variance of $Z^{(t+1)}$ in regions where it is clearly above the target does not matter, but the variance of the process where improvement is uncertain should be reduced.

The ideal sampling criterion for the optimization problem at hand, the variance of the feasible improvement (Equation~\eqref{eq-VFItp1}), is the variance of the product of the improvement times a feasibility with confidence.
The {criterion} $S$ bears some resemblance but is not equivalent to the variance of the feasible improvement. It is a product of variances, which is not equivalent to the variance of a product\footnote{
Let $U$ and $V$ be two random variables, $U \independ V$, $\var(UV) = \mean U^2 \mean V^2 - (\mean U)^2(\mean V)^2$. \\
$\var U\, \var V = (\mean U^2 - (\mean U)^2) (\mean V^2 - (\mean V)^2)$ $= \mean U^2 \mean V^2 - (\mean U)^2 (\mean V^2 - (\mean V)^2) - \mean U^2 (\mean V)^2 $. The variance of the product is the product of the variances only if $\mean U = \mean V = 0$ when $\var(UV) = \var U\, \var V$ $= \mean U^2 \mean V^2$.
}. Furthermore, the second term in the product for $S$ (Equation \eqref{SC}) is an averaged variance of feasibility $\indic_{\cap_{i=1}^l\{G_i^{(t+1)}(\xtarg,\vec{u}) \leq 0 \}}$ as opposed to a variance of feasibility with $1-\alpha$ confidence, $\indic_{C^{(t+1)}(\xtarg)\le 0}$. 

So, we choose the best candidate uncertainty $\vec{u}_{t+1}$ as 
 \begin{equation}
\vec{u}_{t+1}= \arg \min_{\tilde{u} \in  \mathcal{S_U}} {S(\xtarg,\tilde{u})}.
\label{eq-utplus1}
\end{equation} 
The calculation of both terms of $S$ is now presented.

%%%%%%%%%%%%%%%%%%%%%%%%%%%%%%%%%%%%%%%%%%%%%%%%%%%%%%%%
%%%%%%%%%%%%%%%%%%%%%%%%%%%%%%%%%%%%%%%%%%%%%%%%%%%%%%%%

\subsection*{Calculation of $\var\big( I^{t+1}(\xtarg)\big)$}

The expression of $\var\left( I^{t+1}(\xtarg)\right)$ given a new observation at point $(\tilde{\vect{x}},\tilde{\vec{u}})$  is 
\begin{equation}
\begin{split}
\var\left( I^{t+1}(\xtarg)\right)=&~\var\left(\big(z_{\min}^{\text{feas}} - Z^{(t+1)}(\xtarg))^+\right)\\ 
=~&\var\left(\big(z_{\min}^{\text{feas}} - Z(\xtarg)\big)^+| F(D^{(t)})=f^{(t)}, F(\tilde{\vect{x}}, \tilde{u})=f(\tilde{\vect{x}}, \tilde{u})\right)\\
\end{split}
\label{firstterm}
\end{equation}
The expression of the expected improvement in terms of PDF and CDF of Gaussian distribution  is well-known. 
%\chrisdelete{The following gives the formula for the variance (see \ref{proofVI} for the proof).} 
{The closed-form expression of the \emph{Variance of the Improvement} stated in Proposition~\ref{VI} 
can be obtained using the expression of the generalized expected-improvement criterion introduced in \cite{schonlau1998global}. 
For the convenience of the readers, we provide a complete proof in Appendix~\ref{proofVI}.} 

\begin{prop}[The \emph{Variance of the Improvement}]
\begin{equation*}
\var\left(I^{(s)}(\vec{x})\right)= \EI^{(s)}(\vec{x}) \big(z_{\min}^\text{feas} - m_Z^{(s)}(\vec{x}) - \EI^{(s)}(\vec{x})\big) + (\sigma_Z^{(s)})^2(\vec{x})  \Phi\bigg(\frac{z_{\min}^\text{feas} - m_Z^{(s)}(\vec{x})}{\sigma_Z^{(s)}(\vec{x})}\bigg),
\end{equation*}
where $\sigma_Z^{(s)}(\vec{x}) = \sqrt{k_Z^{(s)}(\vec{x},\vec{x})}$.
\label{VI}
\end{prop} 
As $f(\tilde{\vect{x}},\vec{\tilde{u}})$ is unknown, we cannot apply Proposition \ref{VI} to Equation \eqref{firstterm}. 
We use that $F^{(t)}(\tilde{\vect{x}},\vec{\tilde{u}}) \sim \mathcal{N}\big(m_F^{(t)}(\tilde{\vect{x}},\vec{\tilde{u}}) , k_F^{(t)}(\tilde{\vect{x}},\vec{\tilde{u}};\tilde{\vect{x}},\vec{\tilde{u}}) \big)$ and the law of total variance
\begin{equation}
\begin{split}
\var\left( I^{t+1}(\xtarg)\right)=~& \mean\left[ \var\left(\big(z_{\min}^{\text{feas}} - Z(\xtarg)\big)^+ | F(D^{(t)})=f^{(t)}, F(\tilde{\vect{x}}, \tilde{u}) \right)\right]\\
&+ \var\left[ \mean\left(\big(z_{\min}^{\text{feas}} - Z(\xtarg)\big)^+ | F(D^{(t)})=f^{(t)}, F(\tilde{\vect{x}}, \tilde{u}) \right)\right].
\end{split}
\label{firstterm2}
\end{equation}
To compute Equation~\eqref{firstterm2}, notice that the inside of the brackets have closed-form expressions in terms of $m_Z^{(t+1)}$ and $\sigma_Z^{(t+1)}$, which are given by Proposition~\ref{VI} and Equation~\eqref{eq:EI} for the first and second brackets, respectively. 
The external $\mean$ and $\var$ only concern $m_Z^{(t+1)}(\xtarg)$ whose randomness comes from $F(\tilde{\vect{x}},\tilde{u})$, $\sigma_Z^{(t+1)}(\xtarg)$ being not dependent on the evaluation of the function. It is proved in Appendix~\ref{miseajour} that $m_Z^{(t+1)}(\xtarg)$ follows
\begin{equation}
m_Z^{(t+1)}(\xtarg) \sim \mathcal{N}\Bigg( m_Z^{(t)}(\xtarg) , \bigg(\frac{\int_{\mathbb{R}^m} k_F^{(t)}(\xtarg,\vec{u};\tilde{\vec{x}},\vec{\tilde{u}}) \rho_\vec{U}(\vec{u}) d\vec{u} }{\sqrt{k_F^{(t)}(\tilde{\vec{x}},\vec{\tilde{u}};\tilde{\vec{x}},\vec{\tilde{u}})}}\bigg)^2 \Bigg). \\
\label{updatemeanZ}
\end{equation}
Finally, the external $\mean$ and $\var$ in Equation~\eqref{firstterm2} are numerically evaluated. Details are given in Section \ref{sec:implemDetails}.

%%%%%%%%%%%%%%%%%%%%%%%%%%%%%%%%%%%%%%%%%%%%%%%%%%%%%%%%
%%%%%%%%%%%%%%%%%%%%%%%%%%%%%%%%%%%%%%%%%%%%%%%%%%%%%%%%

\subsection*{Calculation of $\int_{\mathbb{R}^m} \var\big( \indic_{\cap_{i=1}^l\{G_i^{(t+1)}(\xtarg,\vec{u}) \leq 0 \}} \big) \rho_\vec{U}(\vec{u}) d\vec{u}$}

Regarding the second term, we follow the \emph{Kriging Believer} principle, leading to suppose that $\forall i=1,\dots,l~;~ m^{(t+1)}_{G_i}(\xtarg,\vec{u})= m^{(t)}_{G_i}(\xtarg,\vec{u})$. 
Under the hypothesis of independence of the constraint GPs, we have 
\begin{equation*}
\int_{\mathbb{R}^m} \var\big( \indic_{\cap_{i=1}^l\{G_i^{(t+1)}(\xtarg,\vec{u}) \leq 0 \}} \big) \rho_\vec{U}(\vec{u}) d\vec{u} = \int_{\mathbb{R}^m} p(\vec{u})(1-p(\vec{u}))\rho_\vec{U}(\vec{u}) d\vec{u},
\end{equation*}
where 
\begin{equation*}
p(\vec{u})=\prod \limits_{i=1}^{l} \Phi\bigg(\frac{-m^{(t)}_{G_i}(\xtarg,\vec{u})}{\sqrt{k^{(t+1)}_{G_i}(\xtarg,\vec{u},\xtarg,\vec{u})}}\bigg),
\end{equation*}
and $k^{(t+1)}_{G_i}(\xtarg,\vec{u},\xtarg,\vec{u}) = k_{G_i}^{(t)}(\xtarg,\vec{u},\xtarg,\vec{u}) - \frac{(k_{G_i}^{(t)}(\xtarg,\vec{u};\tilde{\vec{x}},\vec{\tilde{u}}) )^2} {k_{G_i}^{(t)}(\tilde{\vec{x}},\vec{\tilde{u}};\tilde{\vec{x}},\vec{\tilde{u}})}$ (cf. \cite{chevalier2014corrected}).
%\celi{mettre une ref pour expliquer d'où ça vient}.\rodo{done}
Further details about the numerical estimation of the above integral are given in Section~\ref{sec:implemDetails}.

The steps of the proposed methodology are summarized in Algorithm \ref{algOur}, called \EFISUR for 
Expected Feasible Improvement with Stepwise Uncertainty Reduction sampling.
\begin{algorithm}
\caption{: \EFISUR \small{(Expected Feasible Improvement with Stepwise Uncertainty Reduction sampling)}}
\begin{algorithmic}
\STATE Create an initial Design of Experiments (\DOE) of size $t$ in the joint space and calculate simulator responses: \\$D^{(t)}=\{(\vec{x}_i,\vec{u}_i)~,~i=1,\ldots,t\}$, and associated $f^{(t)}$ and $g_i^{(t)}$
\WHILE{$t~\le $ maximum budget} 
\STATE Create the GPs of the objective and the constraints in the joint space:  $F^{(t)}$ and $(G_i^{(t)})_{i=1}^{l}$
\STATE Calculate the GP of the mean objective, $Z^{(t)}$, in the search space $\mathcal{S_X}$
\STATE \textbf{Optimize} \EFI to define $\xtarg = \arg \max \limits_{\vec{x} \in \mathcal{S_X}} \EFI^{(t)}(\vec{x})$
\quad (Eq.~(\ref{problemEFI}))
\STATE Set $\vec{x}_{t+1} = \xtarg$
\STATE \textbf{Sample} the next uncertain point by solving \\
%\celiadd{$\tilde{u} \rightarrow S(\vec{x_{targ}},\tilde{u})$ } to choose \vec{u_{t+1}},
$\vec{u}_{t+1}= \arg \min_{\tilde{u} \in  \mathcal{S_U}} {S(\xtarg,\tilde{u})}$ \quad (Eq.~(\ref{eq-utplus1}))
\STATE Calculate simulator responses at the next point $(\vec{x}_{t+1},\vec{u}_{t+1})$ 
\STATE Update the \DOE:
$D^{(t+1)} = D^{(t)}\cup (\vec{x}_{t+1},\vec{u}_{t+1})$ , $f^{(t+1)} = f^{(t)} \cup f(\vec{x}_{t+1},\vec{u}_{t+1})$, \\
$g_i^{(t+1)} = g_i^{(t)} \cup g_i(\vec{x}_{t+1},\vec{u}_{t+1})~,~i=1,\ldots,l$ , $t \leftarrow t+1$
\ENDWHILE
\end{algorithmic}
\label{algOur}
\end{algorithm}

%%%%%%%%%%%%%%%%%%%%%%%%%%%%%%%%%%%%%%%%%%%%%%%%%%%%%%%%
%%%%%%%%%%%%%%%%%%%%%%%%%%%%%%%%%%%%%%%%%%%%%%%%%%%%%%%%

\section{Numerical experiments}
\label{sec:numerical}
In this section the performance of the \EFISUR method is studied first on an analytical test case, and then on an industrial application. 
The results are compared to two alternative procedures which are described below.

The code and data generated or used during the current study are available in the first author's GitHub repository \cite{redaGitHubEFISUR}.

\subsection{Competing algorithms}
Two algorithms serve as bases for comparison for \EFISUR. 
{To the best of our knowledge, although inspired by previous works, these algorithms are new.}
First, the \EFIRand algorithm is identical to \EFISUR to the exception of the point $\vec{u}_{t+1}$ 
which is simply sampled from its distribution. 
This \EFIRand algorithm will be useful to judge the usefulness of the sampling criterion $S$ in \EFISUR.
\begin{algorithm}[H]
\caption{: \EFIRand Expected Feasible Improvement with random sampling}
\begin{algorithmic}
\STATE Create an initial \DOE of size $t$ in the joint space and calculate simulator responses: \\$D^{(t)}=\{(\vec{x}_i,\vec{u}_i)~,~i=1,\ldots,t\}$, and associated $f^{(t)}$ and $g_i^{(t)}$
%\WHILE{stopping criterion not met}
\WHILE{$t \le$ maximum budget}
\STATE Create the GPs of the objective and the constraints in the joint space: $F^{(t)}$ and $(G_i^{(t)})_{i=1}^{l}$
\STATE Calculate the GP of the mean objective, $Z^{(t)}$, in the search space $\mathcal{S_X}$
\STATE \textbf{Optimize} \EFI to define $\vec{x}_{t+1} = \arg \max \limits_{\vec{x} \in \mathcal{S_X}} \EFI^{(t)}(\vec{x})$ \quad (Eq.~(\ref{problemEFI}))
\STATE \textbf{Sample} the next uncertain point randomly, $\vec{u}_{t+1} \sim \rho_{\vec{U}}$
\STATE Calculate simulator responses at the next point $(\vec{x}_{t+1},\vec{u}_{t+1})$ 
\STATE Update the \DOE:
$D^{(t+1)} = D^{(t)}\cup (\vec{x}_{t+1},\vec{u}_{t+1})$ , $f^{(t+1)} = f^{(t)} \cup f(\vec{x}_{t+1},\vec{u}_{t+1})$, \\
$g_i^{(t+1)} = g_i^{(t)} \cup g_i(\vec{x}_{t+1},\vec{u}_{t+1})~,~i=1,\ldots,l$ , $t \leftarrow t+1$
\ENDWHILE
\end{algorithmic}
\label{algRand}
\end{algorithm}

The other algorithm {to} which \EFISUR will be compared  uses the quantile as an alternative way of measuring the failure probability.
{
For the first constraint, the quantile writes,
\begin{equation*}
\mathbb{P}(g_1(\vec{x},\vec{U}) \leq 0) \geq 1 - \alpha \Longleftrightarrow q_{1 - \alpha}(g_1(\vec{x},\vec{U})) \leq 0.
\end{equation*}}

{When dealing with several constraints, $\mathbb{P}(g(\vec{x},\vec{U}) \leq 0)$  is then replaced by individuals constraints with quantile level $1- \alpha/l$.
The quantile of the constraint $i=1,\ldots,l$  is approximated using the predictive mean of the GP model $G^{(t)}$,
\begin{equation}
\label{eq:quantileMean}
q_{1-\alpha/l}(g_i(\vec{x},\vec{U})) \approx q_{1-\alpha/l}(m_{G_i}^{(t)}(\vec{x},\vec{U})).
\end{equation}
Further implementation details about the quantile estimation are given in Section \ref{sec:implemDetails}.
}
%\celidelete{and the empirical quantile obtained from 
%$M$ realizations of the uncertainties, $\{\vec{u}_1,\dots,\vec{u}_M\}$, \celi{Attention ici on mélange méthodo et approximation numérique}
%\begin{equation*}
%q_{1-\alpha}(g_i(\vec{x},\vec{U})) \approx q_{1-\alpha}(m_{G_i}^{(t)}(\vec{x},\vec{U})) \approx m_{G_i}^{(t)}(\vec{x},\vec{U})_{(\ceil*{M(1-\alpha)})},
%\end{equation*}
%where we consider the order statistic associated to the sample $\{m_{G_i}^{(t)}(\vec{x},\vec{u}_i) \}_{i=1}^{M}$,
%\begin{equation*}
%m_{G_i}^{(t)}(\vec{x},\vec{U})_{(1)} \leq \dots \leq m_{G_i}^{(t)}(\vec{x},\vec{U})_{(M)}.
%\end{equation*}}

{
In order to choose the next point in the design space, $\vec{x}_{t+1}$, this last competing methodology 
maximizes the expected improvement under constraints about the empirical quantiles:
\begin{equation}
\begin{split}
\vec{x}_{t+1} & = \arg\max \limits_{\vec{x} \in \mathcal{S_X}} \EI^{(t)}(\vec{x}) \\
& \text{ such that } \forall i \in \{1,\dots,l\},~q_{1-\alpha/l}(m_{G_i}^{(t)}(\vec{x},\vec{U})) \leq 0 .
\end{split}
\label{eq:EIconstr}
\end{equation}
}
The sampling in the uncertain space is based on the \textit{deviation number} developed in \cite{echard2011ak,fauriat2014ak}
for the Active Kriging Monte Carlo simulation technique. In these works, the uncertainty that is most likely to improve the GP is the one that minimizes the following \DN (Deviation Number) function,
\begin{equation}\label{eq:Ui}
{\DN}_i(\vec{u}) = \frac{ |m_{G_i^{(t)}}(\mathbf{x}_{t+1},\mathbf{u}) | }{\sigma^{(t)}_{G_i}(\vec{x}_{t+1},\vec{u})}.
\end{equation}
The points that have a low deviation number are either close to the constraint threshold (null in Equation \eqref{eq:Ui}), 
or they have a high GP variance. 
To handle multiple constraints, the constraint with the minimum value of \DN is selected (as in \cite{moustapha2017quantile}), 
\begin{equation}\label{eq:minUi}
\DN_c(\vec{u}) = \min \limits_{i=1,\dots,l} \DN_i(\vec{u}).
\end{equation}
The whole algorithm is called \CEIDev for Constrained \EI plus Deviation Number.
The \CEIDev algorithm is an alternative to \EFISUR for handling chance constraints in the augmented space. 
During the optimization step for $\vec{x}_{t+1}$, the constraints are handled explicitely through an ancilliary constrained optimization algorithm, as opposed to the \EFI that aggregates them in a single objective. 
However, the reliability is defined independently for each constraint, which is not equivalent to the real problem given in Equation~(\ref{problem}).
The sampling step (the \DN minimization) accounts only for the constraints satisfaction and not for the objective function. 
The \CEIDev algorithm bears some resemblance to the approach described in \cite{moustapha2016quantile}: in both cases, the reliability constraint is estimated with a kriging model used to estimate quantiles, and sampling occurs through the minimization of the deviation. 
The generalization of EGO to constrained problems by approximating the constraints through kriging models and keeping them separated from the objective, as it is done in Equation~(\ref{eq:EIconstr}), can be found in other papers, e.g., \cite{bartoli2019adaptive}.
However, to the authors' knowledge, \CEIDev integrates these techniques within an EGO-like algorithm in an original manner. 
\begin{algorithm}[H]
\caption{: \CEIDev}
\label{DN}
\begin{algorithmic}
\STATE Create an initial \DOE of size $t$ in the joint space and calculate simulator responses: \\$D^{(t)}=\{(\vec{x}_i,\vec{u}_i)~,~i=1,\ldots,t\}$, and associated $f^{(t)}$ and $g_i^{(t)}$
\WHILE{$t \le$ maximum budget}
\STATE Create the GPs of the objective and the constraints in the joint space: $F^{(t)}$ and $(G_i^{(t)})_{i=1}^{l}$
\STATE Calculate the GP of the mean objective, $Z^{(t)}$, in the search space $\mathcal{S_X}$
\STATE \textbf{Optimize} the expected improvement under quantile constraints to determine the next iterate 
\begin{equation*}
\vec{x}_{t+1} = \arg\max \limits_{\vec{x} \in \mathcal{S_X}} \EI^{(t)}(\vec{x}) 
\end{equation*}
{
\begin{equation*}
\text{such that } \forall i \in \{1,\dots,l\},~q_{1-\alpha/l}(m_{G_i}^{(t)}(\vec{x},\vec{U})) \leq 0, \quad \quad (Eq.~(\ref{eq:EIconstr}))
%\label{eq:EIconstr}
\end{equation*}
}
\STATE \textbf{Sample} the next uncertainty by minimizing the deviation number, 
\begin{equation*}
\vec{u}_{t+1} = \text{arg} \min \limits_{\mathbf{u}} \DN_c(\vec{u})  \quad \quad (Eq.~(\ref{eq:minUi}))
%\label{eq:usampleDN}
\end{equation*}
\STATE Calculate simulator responses at the next point $(\vec{x}_{t+1},\vec{u}_{t+1})$ 
\STATE Update the \DOE:
$D^{(t+1)} = D^{(t)}\cup (\vec{x}_{t+1},\vec{u}_{t+1})$ , $f^{(t+1)} = f^{(t)} \cup f(\vec{x}_{t+1},\vec{u}_{t+1})$, \\
$g_i^{(t+1)} = g_i^{(t)} \cup g_i(\vec{x}_{t+1},\vec{u}_{t+1})~,~i=1,\ldots,l$ , $t \leftarrow t+1$
\ENDWHILE
\end{algorithmic}
\end{algorithm}

\subsection{Implementation details}
\label{sec:implemDetails}

Two strategies are adopted depending on the numerical cost of the integrals.

\subsubsection*{Common Random Numbers for \vec{u} samples}
%\chris{attention a priori $\vec{U}$ n'est pas uniforme or dans ce paragraphe les formules que l'on donne sont écrites pour des uniformes. deux possibilités soit on dit qu'à partir de maintenant on considère que des uniformes sinon il faudrait faire  mettre à la place de $u_j$ $F(u_j)$ mais il faut supposser que les composantes de U sont indépendantes ce que nous n'avons pas fait.}\rodo{Dans la mesure où les $u_i$ sont tirés suivant leur loi $\rho_U$, les formules me paraissent correctes. Par contre, oui, les tirages $u_i$ sont indépendants}

%\chris{Mon problème est qu'on tire avec une suite de Sobol qui approche une distribution uniforme  sur $\mathcal{S}_U$ qui doit être plus ou moins du  type $[0,1]^m$ et quand je parle d'indépendance ce n'est pas entre les réalisations de $\mathbf{u_i}$ mais entre les composantes d'une réalisation de type $\mathbf{u_i}$} \reda{j'ai enlevé Sobol sequence et mis quantization techniques.}

All three algorithms include Monte Carlo simulations with respect to \vec{u} samples. 
The efficiency of all three algorithms is enhanced by a common random numbers (CRN) technique. 
This means that the same seed is used to generate all random variables throughout the optimization. 
In particular, the same realizations of the uncertain variables $\{\vec{u}_1,\dots,\vec{u}_M\}$, obtained by a 
%\redadelete{Sobol sequence} 
{quantization technique \cite{pages2015introduction}}, are considered in all iterations. 
The CRN produces more stable optimizations and reduces the variance of the estimated probabilities since the induced error is consistent within different designs. 
There is however a bias in the Monte Carlo estimations, which we keep small here by choosing relatively large Monte Carlo number of simulations. 

%\celidelete{As it was seen in Section~\ref{sec-xtarg}, the \EFI acquisition criterion requires 
%the calculation of the quantity $\mathbb{P}(C^{(t)}(\vec{.}) \leq 0)$ 
%which is approximated with $N$ realizations of the GPs:
%\begin{equation}
%\mathbb{P}(C^{(t)}(\mathbf{x}) \leq 0) \approx \frac{1}{N} \sum \limits_{k=1}^{N} \indic_{\big(1 - \alpha - \frac{1}{M} \sum \limits_{j=1}^{M}\indic_{\big(G_{i}^{(t)}(\mathbf{x},\mathbf{u_j},\omega_k) \leq 0,~ i=1,\dots,l \big)} \leq 0\big)}.
%%\label{eq:constrNumEval}
%\end{equation}}

More precisely, the \EFISUR algorithm uses the 
%\redadelete{Sobol sequence} 
{CRN} at different steps:
\begin{itemize}
\item In the \EFI formula, the calculation of the quantity $\mathbb{P}(C^{(t)}(\mathbf{x}) \leq 0)$ is approximated by 
\begin{equation}
\mathbb{P}(C^{(t)}(\mathbf{x}) \leq 0) \approx \frac{1}{N} \sum \limits_{k=1}^{N} \indic_{\big(1 - \alpha - \frac{1}{M} \sum \limits_{j=1}^{M}\indic_{\big(G_{i}^{(t)}(\mathbf{x},\mathbf{u_j},\omega_k) \leq 0,~ i=1,\dots,l \big)} \leq 0\big)}
\label{eq:constrNumEval}
\end{equation}
where $N$ realizations of the GPs are needed.
\item In the $z_{\min}^{\text{feas}}$ formula,  the calculation of $\mean[ C^{(t)}(\vec{x}) ]$ is approximated by :
\begin{equation*}
\mean[ C^{(t)}(\vec{x}) ] \approx 1 - \alpha - \frac{1}{M} \sum \limits_{j=1}^{M} \prod \limits_{i=1}^{l} \Phi\left(\frac{-m_{G_i}^{(t)}(\vec{x},\vec{u_j})}{\sigma_{G_i}^{(t)}(\vec{x},\vec{u_j})}\right) .
\end{equation*}
\item In the second term of the sampling criterion, $\int_{\mathbb{R}^m} p(\vec{u})(1-p(\vec{u}))\rho_\vec{U}(\vec{u}) d\vec{u}$ is approximated by $\frac{1}{M} \sum \limits_{j=1}^{M}  p(\vec{u_j})(1-p(\vec{u_j}))$.
\end{itemize}

The experiments reported in this article have as default $N=1000$ trajectories of the GPs and $M=300$ common random numbers.

The \EFIRand algorithm is identical to \EFISUR to the exception of the sampling of the next uncertain point which is random, hence it has a lower computational complexity.

Concerning the \CEIDev algorithm, the quantiles making the constraints, $q_{1-\alpha}(m_{G_i}^{(t)}(\vec{x},\vec{U})) \leq 0$, are approximated by the corresponding order statistic 
%$m_{G_i}^{(t)}(\vec{x},\vec{U})_{(\ceil*{M(1-\alpha)})}$  
associated to the sample $\{m_{G_i}^{(t)}(\vec{x},\vec{u}_j) \}_{j=1}^{M}$. 
For sharper comparisons, we use the same seed (M = 300 common random numbers) to estimate the quantiles.

{
\subsubsection*{Computational complexity} 
{
Remember that $t$ is the iteration number and the number of data points, $t\le \text{budget}$, $M$ is the number of $\mathbf u$ samples and $N$ is the number of GP trajectories in the simulations, $l$ the number of constraint GPs, $d$ and $m$ the dimensions of $\mathbf u$ and $\mathbf x$, respectively.
The memory usage of the GPs grows in $\mathcal O(t^2)$ and is essentially the same for the \EFISUR, \EFIRand and \CEIDev algorithms.
The details of the calculations of the time complexities of the three algorithms are provided in Appendix~\ref{app-timeCompl}.
The time complexities of the \EFISUR and \EFIRand algorithms are identical and of the order of $\max\left[\mathcal{O}(\text{budget}^3) \times (l + 1) \times (d+m) ~,~ d \times N\times\mathcal{O}(M^3)\right]$, where the first term in $\mathcal O(\texttt{budget}^3)$ comes from learning the GPs in the augmented space while the second term in $\mathcal O(M^3)$ comes from calculating the \EFI. 
Learning the GPs is the critical component of \CEIDev which therefore is computationally cheaper than \EFISUR and \EFIRand in most situations where $M>\texttt{budget}$.  
For large budgets however, the cost of GPs construction will dominate that of the \EFI, making all three algorithms equivalent in terms of complexity.
} % end rodoadd
} % end rev

\subsubsection*{Quantization for $m_Z^{(t+1)}$ samples}
The calculation of the first term of the criterion $S$ requires realizations of $m_Z^{(t+1)}$ (see Equation (\ref{firstterm2})). 
To this end, we 
%\redadelete{use a quantization technique \cite{pages2015introduction}. It consists in approximating} 
{approximate} the continuous distribution of $m_Z^{(t+1)}$ (given in Equation~(\ref{updatemeanZ})) by a discrete one. {As $m_Z^{(t+1)}$ follows a Gaussian distribution, we approximate it using the inverse transformation of the Sobol sequence.} 
Thus, the external variance and expectation of Equation (\ref{firstterm2}) are discretized at values representing the distribution of $m_Z^{(t+1)}$. 
In the upcoming experiments, a quantizer of size $20$ is chosen.

\subsubsection*{Internal optimizations}
The \EFISUR algorithm and its two competitors involve several internal optimization problems that must be solved at every iteration. 
Some of these problems are unconstrained: the maximization of \EFI with respect to \vec{x} (\EFISUR and \EFIRand algorithms) 
or the minimization of $S$ or $U_c$ with respect to \vec{u} (\EFISUR and \CEIDev). 
Such internal continuous unconstrained optimization problems are handled with the derivative-free solver \texttt{BOBYQA} \cite{powell2009bobyqa}. 

The Problem~(\ref{eq:EIconstr}) in the algorithm \CEIDev further requires a solver that can handle constraints. It is addressed with the \texttt{COBYLA} program \cite{powellCOBYLA}.

%%%%%%%%%%%%%%%%%%%%%%%%%%%%%%%%%%%%%%%%%%%%%%%%%%%%%%%%
%%%%%%%%%%%%%%%%%%%%%%%%%%%%%%%%%%%%%%%%%%%%%%%%%%%%%%%%

\subsection{Analytical test case}

%The following settings are common to all the problems. Random Latin hypercube is used to generate the initial designs of experiments. Gaussian Process with Matérn 5/2 autocorrelation function and a constant trend is considered as the default surrogate model.
%\paragraph*{Two-dimensional problem} The first problem uses one design variable, $x$ (bounded by the interval $[-1,1]$), and has one uncertain parameter, $u$ (uniformly distributed over the interval $[0.35,0.65]$. The optimization problem consists in minimizing the expectation over $\vec{U}$ of $f$ under one nonlinear limit state function $g$. The problem is stated as follows:
%\begin{equation}
%\begin{split}
%\text{minimize:} &~~\mean_\vec{U}[f(x,\vec{U})]\\
%\text{satisfying:} &~~\mathbb{P}(g(x,\vec{U}) \leq -2) \geq 1- \alpha\\
%\text{where} &~~ f(x,u) = \big(\frac{1}{3} x^4 - 2.1 u^2+ 4\big)u^2 + x u + 4x^2(x^2 -1)\\
%&~~ g(x,u) = \big( 3x^2 + 7ux - 3 \big) \exp(-x^2) \cos(5\pi x^2) -1.2u^2
%\end{split}
%\label{problem1}
%\end{equation}
%By setting the target probability of failure to $\alpha = 0.05$, the optimum of this problem is found at $x^* = 0.1666$.

\begin{figure}[h!]
	\centering
	\includegraphics[scale=0.5, trim = 0cm 6cm 0cm 5cm, clip]{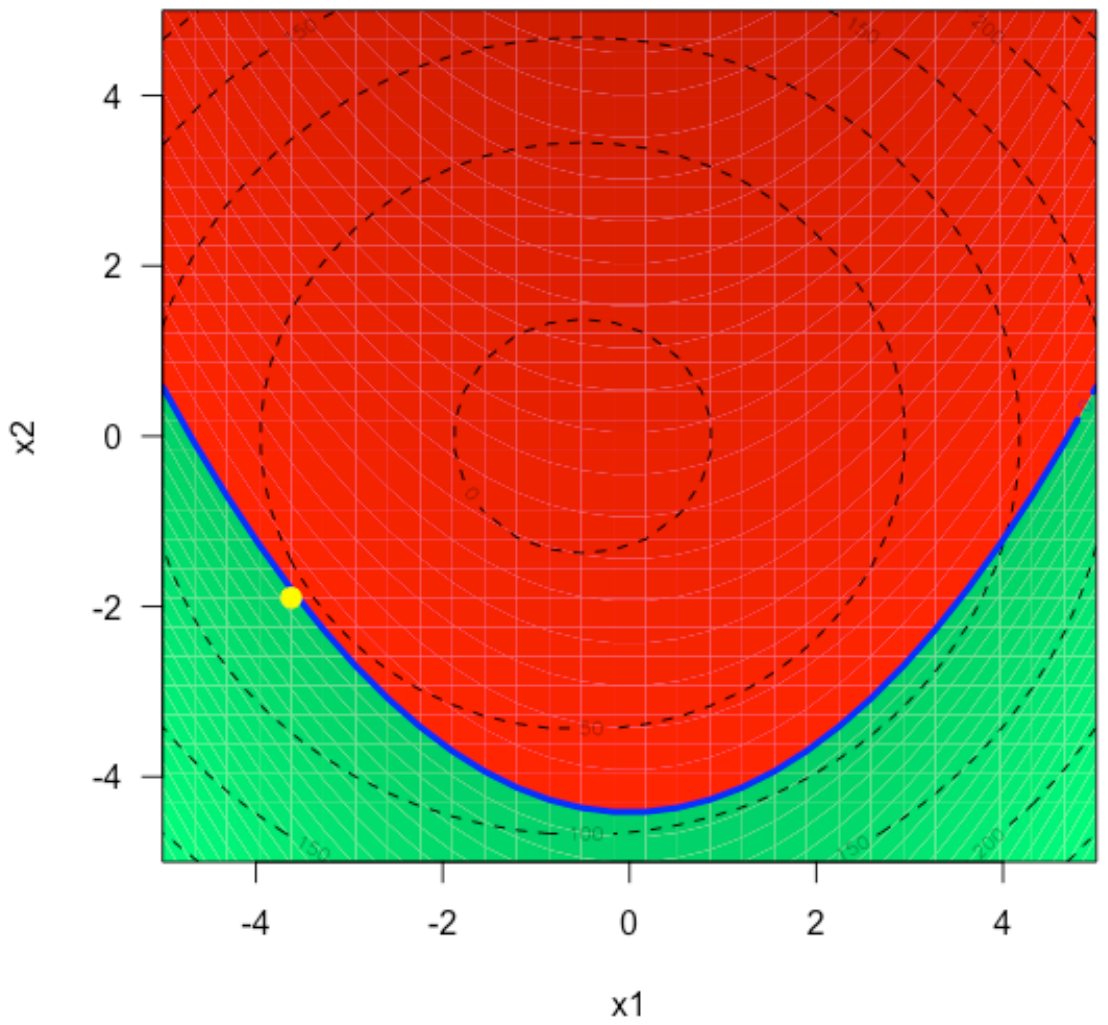}
	\caption{Contour plots of $\mathbb{P}(g(.,\vec{U}) \leq 0)$ and $\mean[f(.,\vec{U})]$: 
	failure and feasible regions in red and green, respectively, the limit-state function in blue, objective function in dashed black lines. The solution is the yellow bullet.}
	\label{test4Dtrue}
	\end{figure}

A first test-case is now considered to compare the three competing algorithms in moderate dimensions. 
{A fully random search, \Random, is added for reference.}
The problem has two design variables, two uncertain parameters and a single reliability constraint: 
\begin{equation*}
\begin{split}
\text{minimize } &~~{\mean[f(\vec{x},\vec{U})]}\\
\text{such that } &~~\mathbb{P}(g(\vec{x},\vec{U}) \leq 0) \geq 1- \alpha\\
\text{where} &~~ f(\vec{x},\vec{u}) = 5(x_1^2+x_2^2) - (u_1^2 + u_2^2) + x_1(u_2-u_1+5) + x_2(u_1-u_2+3)  \\
&~~ g(\vec{x},\vec{u}) = -x_1^2 + 5x_2 - u_1 + u_2^2 -1 \\
\text{with} &~~ \vec{x} \in [-5,5]^2 \\
&~~ \vec{U} \sim \mathcal{U}([-5,5]^2)
\end{split}
\label{eq:problem2}
\end{equation*}
By setting the target probability of failure to $\alpha = 0.05$, the computed reference solution is $\vec{x^*}=(-3.62069,-1.896552)$. 
This reference was found semi-analytically because some of the calculations are manually tractable in the above problem.
Figure \ref{test4Dtrue} shows the contour plots of the functions $\mean[f(.,\vec{U})]$ and $\mathbb{P}(g(.,\vec{U}) \leq 0)$ obtained from a $40\times40$ grid experiments, where at each grid point the expectation and the probability are approximated by a Monte Carlo method over $10^4$ realizations of \vec{U}.
%	\begin{figure}[h!]
%	\centering
%	\includegraphics[scale=0.5]{test4Dtrue.png}
%	\caption{Contour plots of $\mathbb{P}(g(.,\vec{U}) \leq 0)$ and $\mean[f(.,\vec{U})]$: 
%	failure and feasible regions in red and green, respectively, the limit-state function in blue, objective function in dashed black lines. The solution is the yellow bullet.}
%	\label{test4Dtrue}
%	\end{figure}
%	

\begin{figure}[h!]
	\centering
	\includegraphics[scale=0.5, angle = -90]{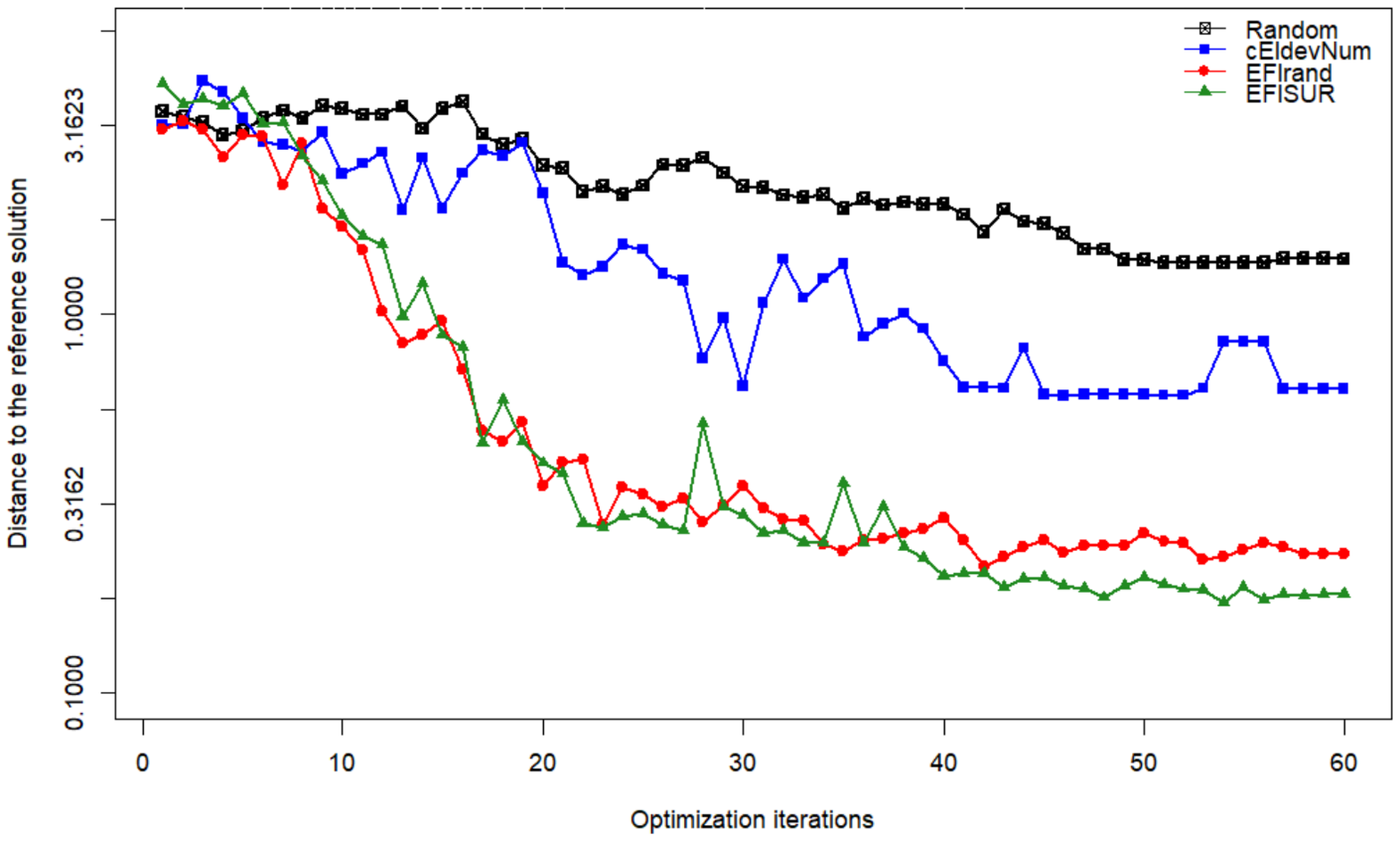}
	\caption{Mean convergence rates (Euclidean distance to the solution) on the analytical test case. 
	The \EFISUR method is plotted with green triangles, \EFIRand with red rounds, \CEIDev with blue squares and \Random with the black crosses. The initial DoE {which precedes} these iterations is not represented.
	}
	\label{fig:4Dmean}
	\end{figure}

	%The 3 algorithms, \EFISUR, \EFIRand and \CEIDev, are compared when solving this test case. 
	To account for the inherent statistical variability of the algorithms, the runs are repeated 30 times for each method. 
	The inital Design of Experiments of each method is a random Latin hypercube of $4+(d+m)=8$ points. 
	An additional budget of 56 iterations\footnote{An iteration encompasses one call to the objective function and one call to all the constraints.} is used as a stopping criterion. 
	The default Gaussian Process has a Matérn 5/2 covariance function and a constant trend. 
	The performance of the various methods is measured by the average Euclidean distance between the optimum given by the method and the true minimum, at each iteration. 
	The distance to the solution{, averaged over the 30 runs,} is plotted in Figure \ref{fig:4Dmean}.
%	\begin{figure}[h!]
%	\centering
%	\includegraphics[scale=0.5]{4Dmean2.png}
%	\caption{Mean convergence rates (Euclidean distance to the solution) on the analytical test case. 
%	The \EFISUR method is plotted with green triangles, \EFIRand with red rounds, \CEIDev with blue squares and \Random with the black diamonds. The initial DoE {which precedes} these iterations is not represented.
%	}
%	\label{fig:4Dmean}
%	\end{figure}
	
\begin{figure}[h!]
	\centering
	\includegraphics[angle = -90, scale=0.6, trim = 4cm 0cm 4cm 0cm, clip]{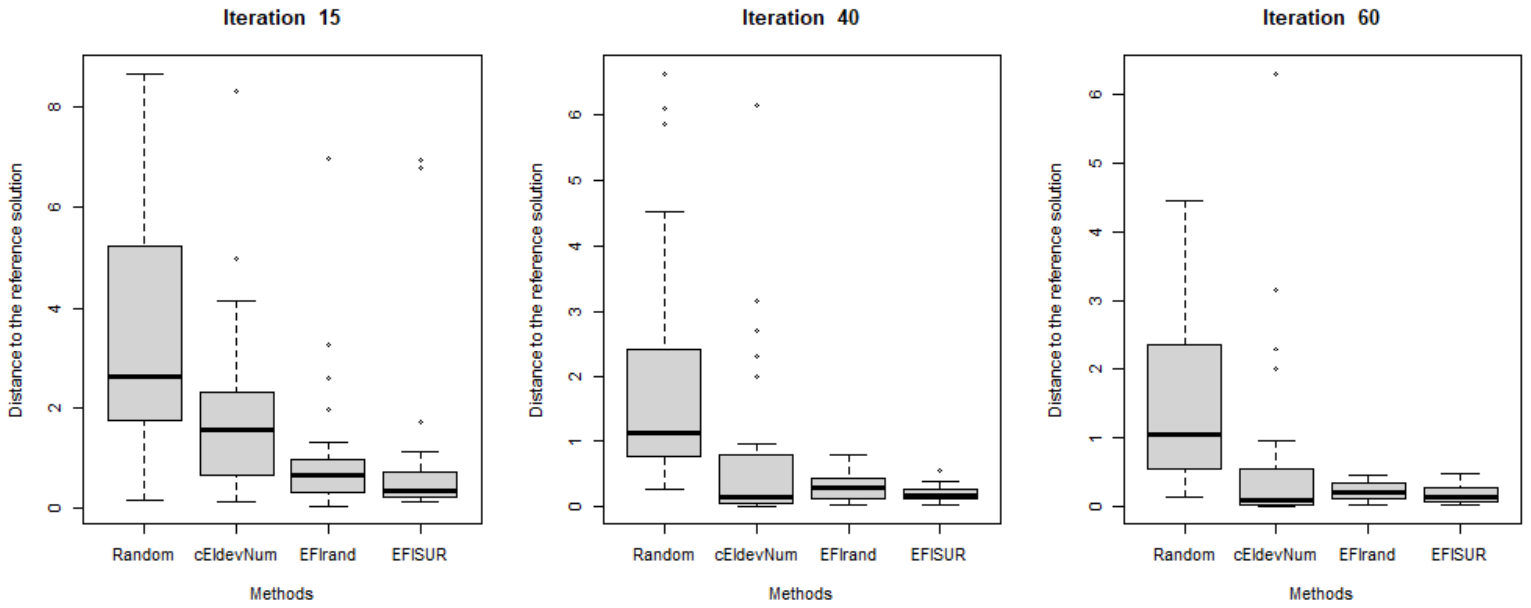}
	\caption{Distance to the reference solution at iteration $15$ (left panel), $40$ (middle panel) and $60$ (right panel) for the three strategies. The boxplots summarize 30 replications of the runs.}
	\label{4Dboxplot}
	\end{figure}

\begin{figure}[h!]
	\centering
	\includegraphics[scale=0.7, trim = 0cm 10cm 0cm 10cm, clip]{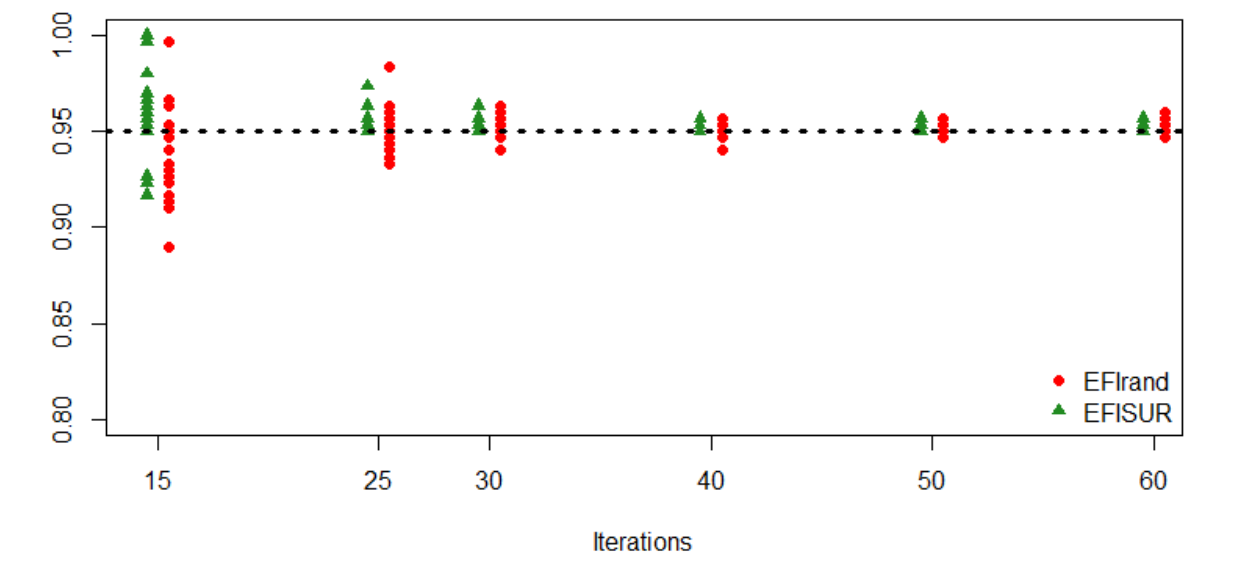}
	\caption{{A posteriori} probability {of satisfying the constraint} at the current ``feasible'' point $z_{\min}^{\text{feas}}$ for different iterations of the \EFISUR and \EFIRand strategies.
}
	\label{4Dprob}
	\end{figure}

	\EFISUR and \EFIRand converge faster to the optimum than \CEIDev. 
	{After 40 iterations, \EFISUR approaches the solution more closely than \EFIRand does.
	A complementary view of the convergence, with dispersions, can be found in Figure \ref{4Dboxplot} which shows the boxplots of the distances to the reference solution at iterations $15$ (left panel), 40 (middle) and $60$ (right). 
	It is observed that \EFISUR leads to an accurate solution 
	from 40 iterations onwards with a small deviation between the runs.
	At all iterations, \CEIDev has a larger {mean} distance to the solution {(but the median is better at 60 iterations)}  and, {mainly,} a larger spread in results. 
Figure \ref{4Dprob} shows the "a posteriori" probability, {calculated by Monte Carlo}, of satisfying the constraint  at the current point {seen as feasible} by the \EFISUR and \EFIRand strategies at different iterations. 
Unlike \EFIRand algorithm, starting from iteration $25$, all the points proposed by the \EFISUR algorithm satisfy the probabilistic constraint. 
Thus, the points proposed by \EFISUR are not only close to the reference solution, but they are also inside the safe region.}
%	\begin{figure}[h!]
%	\centering
%	\includegraphics[scale=0.5, trim = 0cm 10cm 0cm 10cm, clip]{4Dboxplot1.png}
%	\caption{Distance to the reference solution at iteration $15$ (left panel), $40$ (middle panel) and $60$ (right panel) for the three strategies. The boxplots summarize 30 replications of the runs.}
%	\label{4Dboxplot}
%	\end{figure}
	
%	\begin{figure}[h!]
%	\centering
%	\includegraphics[scale=0.7]{4Dprob2.png}
%	\caption{{A posteriori} probability {of satisfying the constraint} at the current ``feasible'' point $z_{\min}^{\text{feas}}$ for different iterations of the \EFISUR and \EFIRand strategies.
%}
%	\label{4Dprob}
%	\end{figure}
\begin{figure}[h!]
	\centering
	\includegraphics[scale=0.75, trim = 0cm 11cm 0cm 10cm, clip]{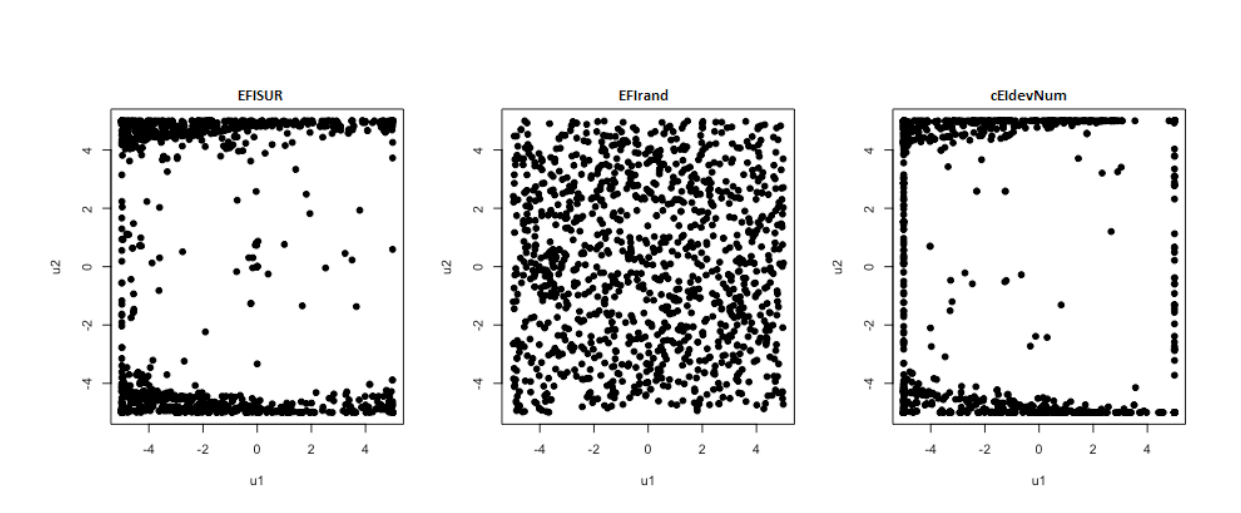}
	\caption{Enrichment in the uncertain space, $\mathcal{S_\vec{U}}$, for the three methods.}
	\label{4DUnc}
	\end{figure}

	Figure \ref{4DUnc} shows the enrichment in the uncertain space $\mathcal{S_\vec{U}}$ for all methods and all runs. 
	It is clearly visible that \EFIRand, in the middle plot, samples the \vec{u}'s randomly.
{The Random search does the same, of course.}
	\EFISUR (left) and \CEIDev (right) both sample large $\lvert u_2 \rvert$'s because they contribute to constraint violation irrespectively of \vec{x} through the ``$+u_2^2$'' term (cf. $g(\vec{x},\vec{u})$ expression above). 
	In addition, \CEIDev samples large values $\lvert u_1 \rvert$'s for varied $u_2$'s because they are on the edge of the domain where the kriging variance is large (hence \DN small). 
	\EFISUR breaks this symmetry and more sparingly tries small (negative) $u_1$'s associated to varied $u_2$'s because its criterion also accounts for low objective function values: 
	in the optimal region, $\vec{x}^* \approx (-3.6,-1.9)$, a small negative $u_1$ provides improvement through the term ``$-x_1 u_1$''.

%	\begin{figure}[h!]
%	\centering
%	\includegraphics[scale=0.7]{4DUnc.png}
%	\caption{Enrichment in the uncertain space, $\mathcal{S_\vec{U}}$, for the three methods.}
%	\label{4DUnc}
%	\end{figure}

	Two partial conclusions can be presumed from this set of experiments. 
	First, the algorithms that rely on the \EFI aggregated criterion to choose the next set of design variables, i.e., \EFIRand and \EFISUR, converge better than \CEIDev and its constrained problem handled through \texttt{COBYLA}.
	Second, \EFISUR provides an additional gain in efficiency thanks to its sampling criterion $S$ which properly mixes the uncertainty about constraint satisfaction and improvement.

	%%%%%%%%%%%%%%%%%%%%%%%%%%%%%%%%%%%%%%%%%%%%%%%%%%%%%%%%
	%%%%%%%%%%%%%%%%%%%%%%%%%%%%%%%%%%%%%%%%%%%%%%%%%%%%%%%%

\subsection{Industrial test case}
We now report the application of the \EFISUR method to an aeronautical test case. 
The NASA rotor 37 is a representative transonic axial-flow compressor that has been used extensively in the computational fluid dynamics (CFD) community to test optimization algorithms and validate CFD codes (see \cite{hirsch2019uncertainty}). 
The optimization of the NASA rotor 37 compressor blade is a challenging test case primarily because of its high dimensionality: 
it has 20 design variables and 7 uncertain parameters. 
As such, to the best of the author’s knowledge, such optimization has never been attempted using global metamodels. 
Furthermore, the design of the NASA rotor 37 compressor blade is highly nonlinear. And, as is common in CFD, each 
evaluation of the optimization criteria involves costly finite elements analyses.
Formally, the optimization problem reads as follows:
	\begin{equation*}
	\begin{split}
	\text{minimize:} &~~{\mean [f(\vec{x},\vec{U})]}\\
	\text{satisfying:} &~~\mathbb{P}(g_i(\vec{x},\vec{U}) \leq 0, \forall i \in \{1,\dots,5\}) \geq 1- \alpha\\
	\text{with} &~~ \alpha = 5\%, \\
	&~~ \vec{x} \in \mathcal{S_X} \quad,\quad \mathcal{S_X} = [0,1]^{20} \subset \mathbb{R}^{20}, \\
	%&~~ \vec{U} \sim \mathcal{U}(\Xi).
	&~~ \vec{U} \sim \mathcal{U}(\mathcal{S_U}), \quad,\quad \mathcal{S_U} = [0,1]^7 \subset \mathbb{R}^{7}.\\
	%&~~ \mathcal{S_U} = [0.05,0.80] \times [4\text{e-7},12\text{e-7}] \times [0.98,1.02] \times [-5.763,5.763] \times\\
	%&~~~~~~~~~~[-0.5,0.5] \times [19.78,20.59] \times [16845,17532].
	\end{split}
	\label{problem3}
	\end{equation*}

\begin{figure}[h!]
	\centering
	\includegraphics[angle = -90,scale=0.5, trim = 1cm 3cm 2cm 1cm, clip]{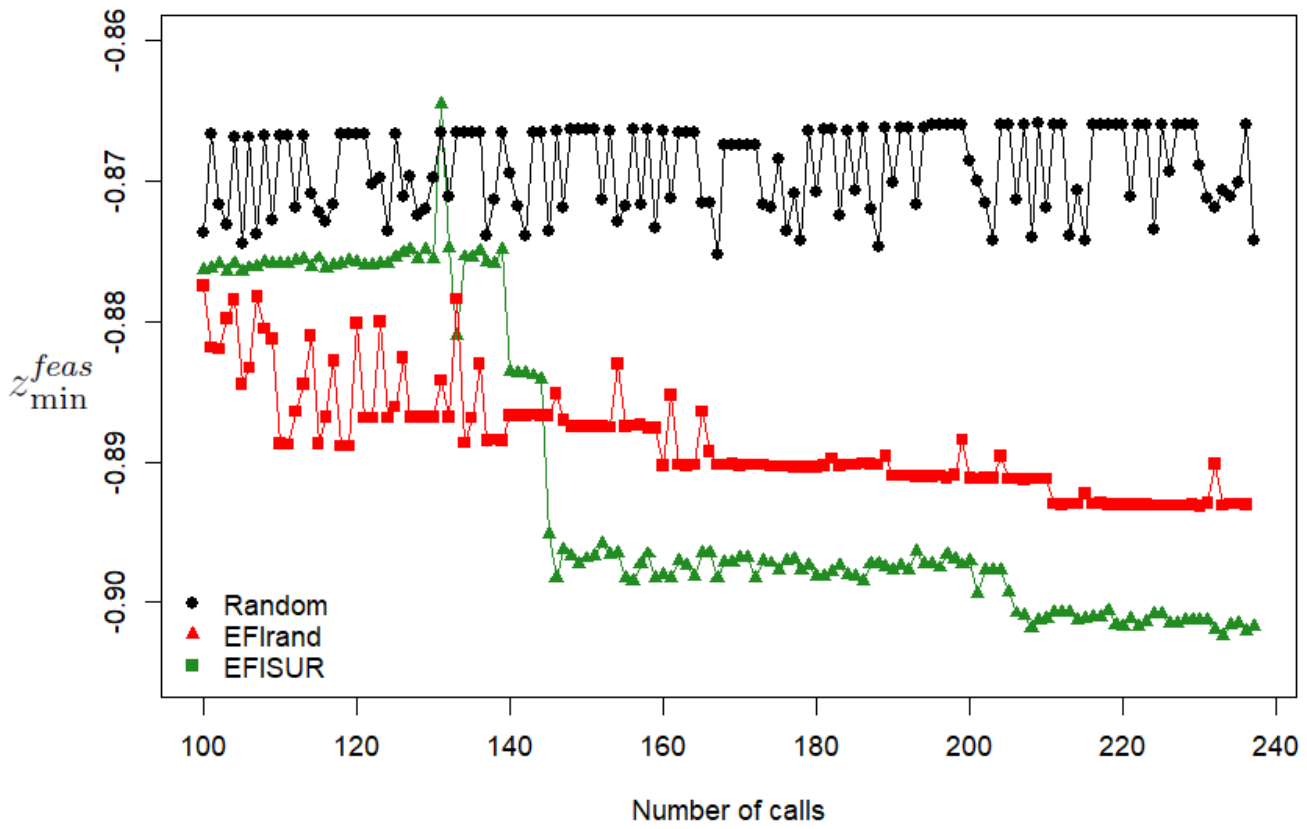}
	\caption{Convergence history of {the average objective function of} the current feasible minimum, $z_{\min}^{\text{feas}}$.}
	\label{Safran}
	\end{figure}

Because of the dimensionality and the numerical cost, the use of surrogate models is the only path to perform such an optimization. {To highlight the importance of carefully sampling in the joint-space during the optimization, we compare our approach to the \EFIRand algorithm}. 
{For comparison purposes, we also carry out a random search.} All the methods are started with a \DOE of $100$ points drawn from an optimal Latin Hypercube Sampling. 
The enrichment then proceeds by adding $137$ points, for a total of $237$ model evaluations. Figure \ref{Safran} shows the convergences of the feasible minimum {in terms of mean objective function. We first note that the \EFIRand and \EFISUR methods very clearly give access to better designs than the random search does}. 
%{We observe a faster convergence of our approach in contrast to \EFIRand. This behavior is clearly attributed to the sampling in the uncertain space, which results in an accurate estimation of the probabilistic constraints and, consequently, the boundary limit.}\rodo{version modifiée ci-dessous}
{We observe that, after a warm-up phase of about 140 analyses, \EFISUR finds better designs than \EFIRand does.} 
{This behavior is attributed to the only difference between \EFISUR  and  \EFIRand, that is the sampling in the uncertain space.} 
%\rododelete{{Properly choosing the random variables $\vec{u}$ results in a more accurate estimation of the probabilistic constraints and, consequently, of the boundary limit.}}\rodo{le choix de \vec{u} mélange la précision dans les contraintes et le progrès de la moyenne de l'objectif. Ou a-t-on une bonne raison de penser que ça se joue principalement sur les contraintes? Je propose la reformulation ci-après}
{Properly choosing the random variables $\vec{u}$ results in a more accurate estimation of the criterion of merit, the feasible improvement, in the target region, which translates into more frequent progress.}

%	\begin{figure}[h!]
%	\centering
%	\includegraphics[scale=0.5]{Safran5.png}
%	\caption{Convergence history of {the average objective function of} the current feasible minimum, $z_{\min}^{\text{feas}}$.}
%	\label{Safran}
%	\end{figure}

\begin{figure}[h!]
	\centering
	\includegraphics[scale=0.45, angle = -90,trim = 0cm 0cm 0cm 0cm, clip ]{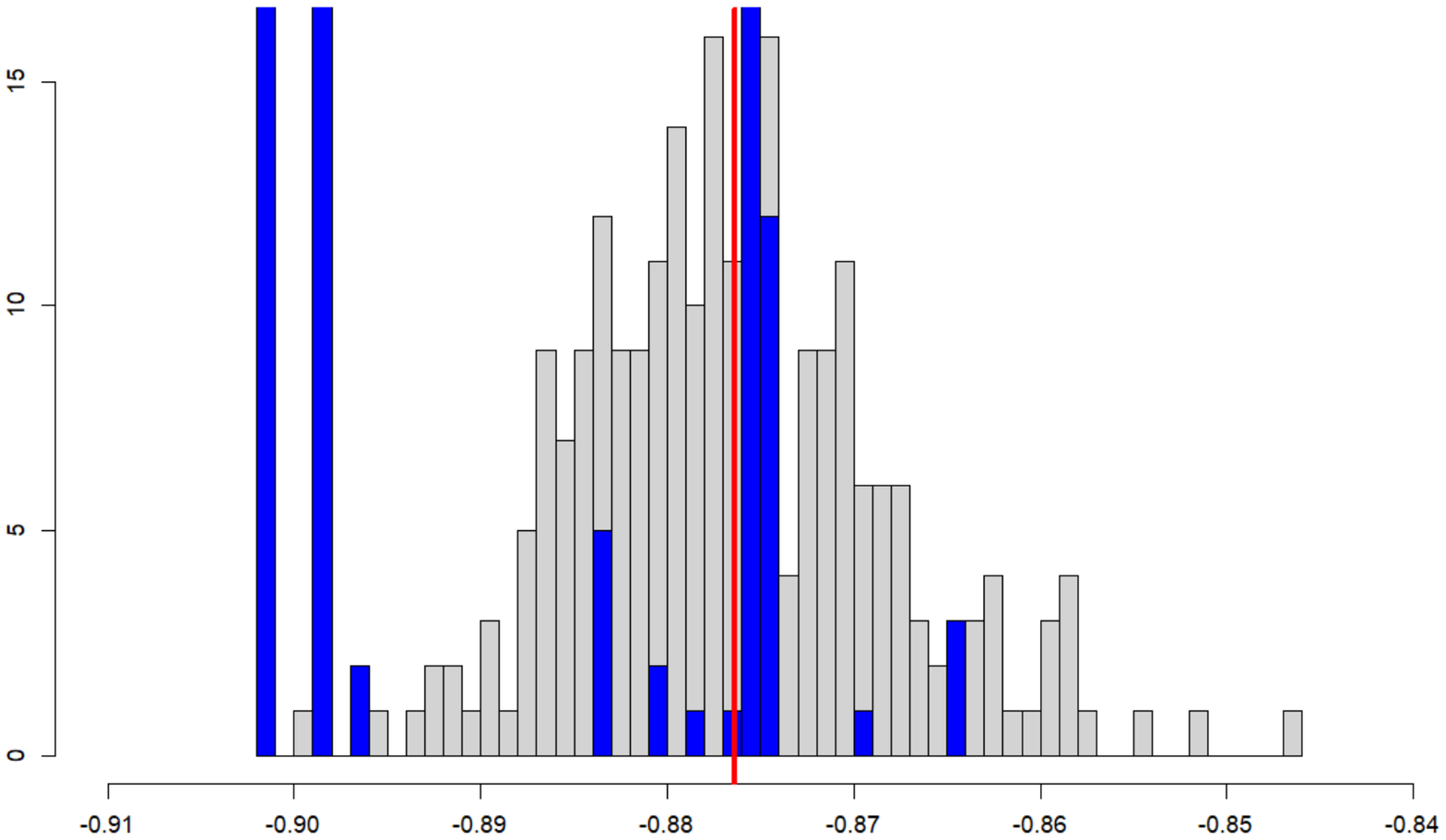}
	\caption{{Histograms of \textit{i)} in grey, the $237$ estimated mean objectives ($m_Z^{(T)}(\vec{x})$) of the initial DoE followed by the optimization, 
\textit{ii)} in blue, the $137$ mean objectives of the best feasible point generated during the optimization. The red line corresponds to the mean objective of the best feasible point at the first iteration.}}
	\label{Histogram}
	\end{figure}

{Figure \ref{Histogram} shows two distributions of the objective: first, in grey, all the estimated mean objective functions $m_Z^{(T)}(\vec{x})$ of the points evaluated during, both, the initial DoE and the optimization; second, in blue, the estimated mean objective of the best feasible point, $z_{\min}^{\text{feas}}$, during the optimization iterations. Note that the best feasible objective has shifted from $-0.876$ at the end of the DoE to $-0.90174$ after the optimization, representing $51.2\%$ of the objective's range. 
All mean objectives shown on the Figure are evaluated with the last, most up-to-date, GPs. 
}
%	\begin{figure}[h!]
%	\centering
%	\includegraphics[scale=0.6]{SafranHist4.png}
%	\caption{{Histograms of \textit{i)} in grey, the $237$ estimated mean objectives ($m_Z^{(T)}(\vec{x})$) of the initial DoE followed by the optimization, 
%\textit{ii)} in blue, the $137$ mean objectives of the best feasible point generated during the optimization.}}
%	\label{Histogram}
%	\end{figure}

\begin{figure}[h!]
	\centering
	\includegraphics[scale=0.60, trim = 1cm 9cm 2cm 10cm, clip ]{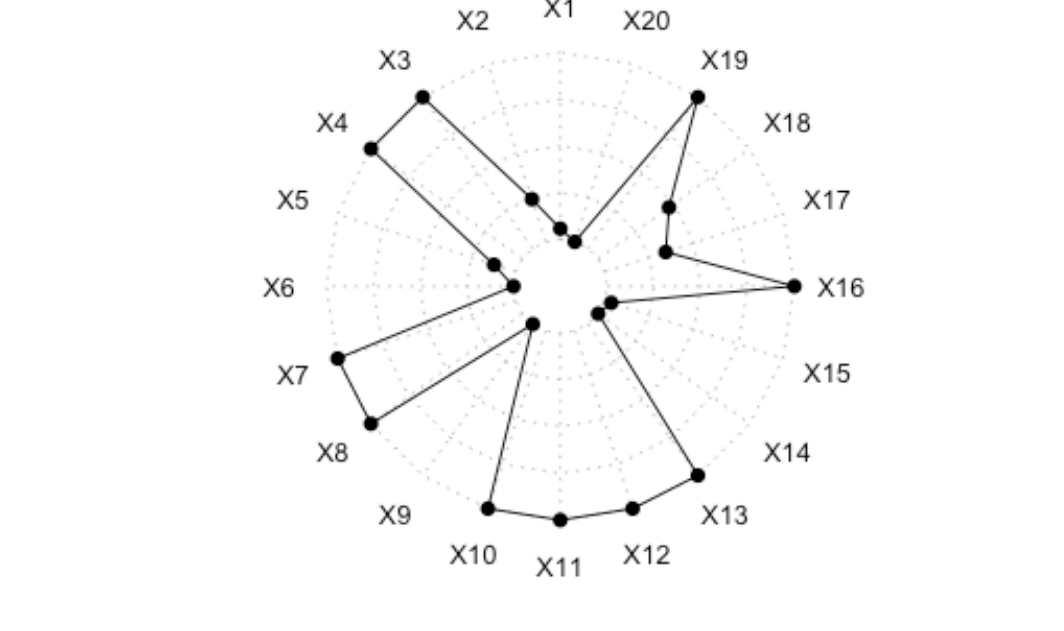}
	\caption{Relative coordinates of the optimal design with respect to their respective lower and upper bounds.}
	\label{RadarDesign}
	\end{figure}

Figure \ref{RadarDesign} shows, at the estimated feasible optimum, the relative value of each design variable with respect to its lower and upper bounds in polar coordinates. 
The largest radii are attributed to the variables which are close to their maximum allowable values, and vice versa. 
%	\begin{figure}[h!]
%	\centering
%	\includegraphics[scale=0.60]{RadarDesign.png}
%	\caption{Relative coordinates of the optimal design with respect to their respective lower and upper bounds.}
%	\label{RadarDesign}
%	\end{figure}

\begin{figure}[h!]
\centering
\includegraphics[scale=0.9, trim = 1cm 11cm 2cm 11cm, clip ]{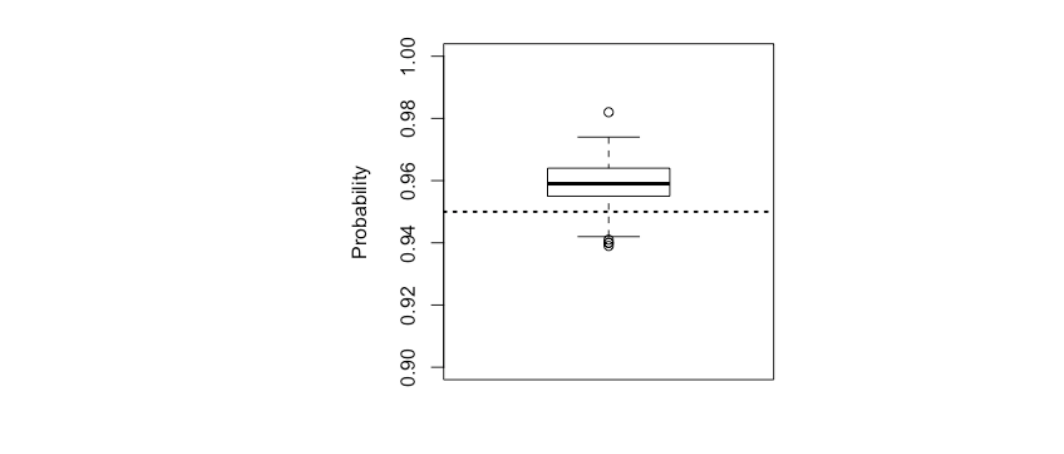}
\caption{Distribution of constraints satisfaction probabilities at the optimal design computed with the final GPs. The boxplot summarizes 500 replications. The dashed line is the lower bound on constraint satisfaction (0.95).
}
\label{BoxplotConst}
\end{figure}

{Because it is generated by \EFISUR,} the above result assumes that the final GPs are accurate enough to correctly predict the probability of constraints satisfaction. 
In order to validate the surrogate model accuracy in the vicinity of the limit-state surface, the calculation of the probability of being
feasible (Equation~(\ref{eq:constrNumEval})) is repeated 500 times with $M=1000$ \vec{u} samples and $N=1000$ trajectories of the GPs.
%a set of $1000$ uncertain parameters \vec{u} is drawn.
%Then, 500 bootstrap replications of $M=500$ parameters are taken from this set and the probability of constraint satisfaction re-estimated at the final design \vec{x} (which involves $N=1000$ trajectories of the final GPs at these points, cf. Equation~(\ref{eq:constrNumEval})). 
Figure \ref{BoxplotConst} provides the statistics of the probability of constraints satisfaction through a boxplot. 
Accounting for the bootstrap standard deviation, the targeted probability (0.95) remains below the confidence interval of the estimated probabilities. Thus the final GP models of the constraints are deemed accurate enough.
%\begin{figure}[h!]
%\centering
%\includegraphics[scale=0.9]{BoxplotConst.png}
%\caption{Distribution of constraints satisfaction probabilities at the optimal design computed with the final GPs. The boxplot summarizes 500 replications. The dashed line is the lower bound on constraint satisfaction (0.95).
%}
%\label{BoxplotConst}
%\end{figure}
It is worth emphasizing that only $237$ calculations of $f$ and the $g_i$'s have been necessary to solve this 27-dimensional optimization problem.

%%%%%%%%%%%%%%%%%%%%%%%%%%%%%%%%%%%%%%%%%%%%%%%%%%%%%%%%
%%%%%%%%%%%%%%%%%%%%%%%%%%%%%%%%%%%%%%%%%%%%%%%%%%%%%%%%

\section{Concluding remarks}
% Rodo : j'ai vérifié le temps et c'est ok: present perfect pour ce qui a été fait (les explications passées du papier) et est relié au présent (de l'écriture de la conclusion), puis présent pour ce qui est toujours valable.
We have proposed a robust Bayesian optimization algorithm to solve computationally intensive chance constrained problems. 
The algorithm, called \EFISUR, carefully models all available information with Gaussian processes 
built in the augmented space of the design variables and uncertain parameters. 
New calls to the objective and constraint functions are based on extrema of an acquisition criterion followed by a sampling criterion. 
The acquisition criterion is an expected feasible improvement that accounts for both the average improvement in objective function and the constraint reliability. 
The associated sampling criterion is {inspired by} the one-step-ahead variance reduction in feasible improvement {and computationally tractable}. 

The article has detailed the analytical expressions of the acquisition and sampling criteria. 

\EFISUR has been compared to two alternative algorithms which differ in the acquisition and sampling criteria. 
The results show a gain in favor of the expected feasible improvement and {the proposed} one-step-ahead variance reduction {criterion}. 
This set of criteria accounts for both the objective function and the constraints, it is opportunistic in the sense that it strives for feasible improvement at the next iteration, and both criteria (\EFI and $S$) are coherent because they both relate to the feasible improvement. 
The sampling criterion follows a principle of maximal uncertainty reduction. 
The applicability of \EFISUR to an industrial test case has also been checked.

From a methodological perspective, further work on the topic might seek to account for the correlation that typically exists between the constraint functions or between the objective and the constraint functions. 
%\rododelete{with the main challenge of having to define the analytical expression for the feasible improvement under this assumption. }\rodo{je ne crois pas que cela change l'expression}\chris{ je suis d'accord si on met juste de la corrélation entre les contraintes mais si on met de la corrélation entre l'objectif et les contraintes je pense que cela change et que tout sera que numérique}\rodo{Ok, c'est vrai, mais je n'en parle pas car je trouve que ça complique trop la conclu}
This would potentially improve the overall Gaussian model of the optimization functions. 
It should also make it possible to assign priorities for evaluating the constraints.

\section*{Acknowledgements}
The first author was supported by the research chair OQUAIDO and the consortium CIROQUO gathering partners in technological research and academia in the development of advanced methods for Computer Experiments. \\
We are very grateful to the three anonymous reviewers, whose comments improved the paper.

\section*{Code and data availability} The code and data generated or used during the current study are available in the first author's GitHub repository \cite{redaGitHubEFISUR}.
%\paragraph{Conflicts of interest:} The authors have no conflicts of interest to declare that are relevant to the content of this article.

%%%%%%%%%%%%%%%%%%%%%%%%%%%%%%%%%%%%%%%%%%%%%%%%%%%%%%%%
%%%%%%%%%%%%%%%%%%%%%%%%%%%%%%%%%%%%%%%%%%%%%%%%%%%%%%%%

\bibliographystyle{plain}
\bibliography{article_EFISUR_SMAI_JCM_2023}

%%%%%%%%%%%%%%%%%%%%%%%%%%%%%%%%%%%%%%%%%%%%%%%%%%%%%%%%
%%%%%%%%%%%%%%%%%%%%%%%%%%%%%%%%%%%%%%%%%%%%%%%%%%%%%%%%
%\newpage
\appendix

\section{Expression of the aggregated variance sampling criterion}
\subsection{The \emph{Variance of the Improvement}}
\label{proofVI}
%\chrisdelete{We now prove the closed-form expression of the \emph{Variance of the Improvement} stated in Proposition~\ref{VI}:}
{The closed-form expression of the \emph{Variance of the Improvement} stated in Proposition~\ref{VI} can be obtained using the expression of generalized expected-improvement criterion introduced  in \cite{schonlau1998global}. 
For the convenience of the reader we still recall the details of the proof.}

\begin{proof}
\begin{equation*}
\begin{split}
VI(\vec{x}) = \var(I(\vec{x}))  =&~ \mean[I^2(\vec{x})] - \mean[I(\vec{x})] ^2 \\
=&~ \int_{-\infty}^{z_{\min} } ( z_{\min} - z)^2 f_{\mathcal{N}(m_Z(\vec{x}),\sigma_Z^2(\vec{x}))}(z) dz - (\EI(\vec{x}))^2,\\
=&~ \int_{-\infty}^{\frac{z_{\min} - m_Z(\vec{x})}{\sigma_Z(\vec{x})}}  (z_{\min} - m_Z(\vec{x}) - \sigma_Z(\vec{x}).{v} )^2 {\phi(v) dv} - (\EI(\vec{x}))^2,\\
=&~ \sigma_Z^2(\vec{x}) \int_{-\infty}^{\frac{z_{\min} - m_Z(\vec{x})}{\sigma_Z(\vec{x})}} {v}^2 \phi({v}) d{v} + (z_{\min} - m_Z(\vec{x}))^2 \int_{-\infty}^{\frac{z_{\min} - m_Z(\vec{x})}{\sigma_Z(\vec{x})}} \phi({v}) d{v} \\
&- 2\sigma_Z(\vec{x}) (z_{\min} - m_Z(\vec{x}))\int_{-\infty}^{\frac{z_{\min} - m_Z(\vec{x})}{\sigma_Z(\vec{x})}} {v} \phi({v}) d{v} - (\EI(\vec{x}))^2, \\
=&~ \sigma_Z^2(\vec{x}) \int_{-\infty}^{\frac{z_{\min} - m_Z(\vec{x})}{\sigma_Z(\vec{x})}} v^2 \phi({v}) d{v} + (z_{\min} - m_Z(\vec{x}))^2 \Phi\bigg(\frac{z_{\min} - m_Z(\vec{x})}{\sigma_Z(\vec{x})}\bigg) \\
&+ 2\sigma_Z(\vec{x}) (z_{\min} - m_Z(\vec{x})) \phi\bigg(\frac{z_{\min} - m_Z(\vec{x})}{\sigma_Z(\vec{x})}\bigg) - (\EI(\vec{x}))^2, \\
=&~ \sigma_Z^2(\vec{x}) \bigg( [-{v}\phi({v}) ]_{-\infty}^{\frac{z_{\min} - m_Z(\vec{x})}{\sigma_Z(\vec{x})}} + \int_{-\infty}^{\frac{z_{\min} - m_Z(\vec{x})}{\sigma_Z(\vec{x})}} \phi({v}) d{v}\bigg) \\
&+ (z_{\min} - m_Z(\vec{x}))^2 \Phi\bigg(\frac{z_{\min} - m_Z(\vec{x})}{\sigma_Z(\vec{x})}\bigg)\\
& + 2\sigma_Z(\vec{x}) (z_{\min} - m_Z(\vec{x})) \phi\bigg(\frac{z_{\min} - m_Z(\vec{x})}{\sigma_Z(\vec{x})}\bigg) - (\EI(\vec{x}))^2, \\
=&~ - \sigma_Z(\vec{x}) (z_{\min} - m_Z(\vec{x})) \phi\bigg(\frac{z_{\min} - m_Z(\vec{x})}{\sigma_Z(\vec{x})}\bigg)\\
&+ \sigma_Z^2(\vec{x}) \Phi\bigg(\frac{z_{\min} - m_Z(\vec{x})}{\sigma_Z(\vec{x})}\bigg) + (z_{\min} - m_Z(\vec{x}))^2 \Phi\bigg(\frac{z_{\min} - m_Z(\vec{x})}{\sigma_Z(\vec{x})}\bigg)\\
&+ 2\sigma_Z(\vec{x}) (z_{\min} - m_Z(\vec{x})) \phi\bigg(\frac{z_{\min} - m_Z(\vec{x})}{\sigma_Z(\vec{x})}\bigg) - (\EI(\vec{x}))^2, \\
=&~  \bigg( \big(z_{\min}  - m_Z(\vec{x})\big)^2 + \sigma_Z^2(\vec{x})\bigg) \Phi\bigg(\frac{z_{\min} - m_Z(\vec{x})}{\sigma_Z(\vec{x})}\bigg)\\
&+ \sigma_Z(\vec{x}) (z_{\min} - m_Z(\vec{x})) \phi\bigg(\frac{z_{\min} - m_Z(\vec{x})}{\sigma_Z(\vec{x})}\bigg) - (\EI(\vec{x}))^2,\\
=&~  \EI(\vec{x}) \big(z_{\min}  - m_Z(\vec{x}) - \EI(\vec{x})\big) + \sigma_Z^2(\vec{x})  \Phi\bigg(\frac{z_{\min} - m_Z(\vec{x})}{\sigma_Z(\vec{x})}\bigg).
\end{split}
\end{equation*}
\end{proof}
Use was made of the expression of the expected improvement, \EI, Equation~(\ref{eq:EI}). This shows that the \emph{Variance of the Improvement} can be expressed in terms of PDF and CDF of $\mathcal{N}(0,1)$.

%%%%%%%%%%%%%%%%%%%%%%%%%%%%%%%%%%%%%%%%%%%%%%%%%%%%%%%%
%%%%%%%%%%%%%%%%%%%%%%%%%%%%%%%%%%%%%%%%%%%%%%%%%%%%%%%%

\subsection{Variance of the Improvement at step $t+1$} \label{miseajour}

Recall the expression of the variance of the improvement given in Equation~(\ref{firstterm2}) that results from the law of total variance,
\begin{equation*}
\begin{split}
\var&\left( I^{(t+1)}(\xtarg)\right)=~ \mean\left[ \var\left(\big(z_{\min}^{\text{feas}} - Z(\xtarg)\big)^+ | F(D^{(t)})=f_t, F(\tilde{\vect{x}}, \tilde{u}) \right)\right]\\
&\hspace{0.8cm}+ \var\left[ \mean\left(\big(z_{\min}^{\text{feas}} - Z(\xtarg)\big)^+ | F(D^{(t)})=f_t, F(\tilde{\vect{x}}, \tilde{u}) \right)\right] \\
&= \mean[ VI^{(t+1)}(\xtarg) | m_Z^{(t+1)}(\xtarg) ] + \var( \EI^{(t+1)}(\xtarg) | m_Z^{(t+1)}(\xtarg) ]).
\end{split}
\end{equation*}
The last line is because the improvement variance and expectation at $t+1$ are closed-form 
expressions of $m_Z^{(t+1)}$ and $\sigma_Z^{(t+1)}$ and only $m_Z^{(t+1)}$ is random through $F(\tilde{\vect{x}}, \tilde{u})$.

It is now proved that the one-step-ahead mean of $F$, $m_Z^{(t+1)}$, is also Gaussian. 
The proof makes use of the one step update formula which is found in \cite{phd_chevalier}:
\begin{equation*}
\left\{
\begin{array}{ll}
&m_F^{(t+1)}(\vec{x},\vec{u}) = m_F^{(t)}(\vec{x},\vec{u}) + \frac{k_F^{(t)}(\vec{x},\vec{u};\tilde{\vec{x}},\vec{\tilde{u}})}{k_F^{(t)}(\tilde{\vec{x}},\vec{\tilde{u}};\tilde{\vec{x}},\vec{\tilde{u}})} \big(  F(\tilde{\vec{x}},\vec{\tilde{u}})  - m_F^{(t)}(\tilde{\vec{x}},\vec{\tilde{u}}) \big), \\
&k_F^{(t+1)}(\vec{x},\vec{u},\vec{x'},\vec{u'}) = k_F^{(t)}(\vec{x},\vec{u},\vec{x'},\vec{u'}) - \frac{k_F^{(t)}(\vec{x},\vec{u};\tilde{\vec{x}},\vec{\tilde{u}}) k_F^{(t)}(\vec{x'},\vec{u'};\tilde{\vec{x}},\vec{\tilde{u}})} {k_F^{(t)}(\tilde{\vec{x}},\vec{\tilde{u}};\tilde{\vec{x}},\vec{\tilde{u}})},
\end{array}
\right.
\end{equation*}
where $((\tilde{\vec{x}},\vec{\tilde{u}});F(\tilde{\vec{x}},\vec{\tilde{u}}))$ is the new, yet still random, observation.
\begin{equation*}
\begin{split}
m_Z^{(t+1)}(\xtarg) &= \int_{\mathbb{R}^m} m_F^{(t+1)}(\xtarg,\vec{u}) \rho_\vec{U}(\vec{u}) d\vec{u},\\
 &= \int_{\mathbb{R}^m} \bigg(m_F^{(t)}(\xtarg,\vec{u}) + \frac{k_F^{(t)}(\xtarg,\vec{u};\tilde{\vec{x}},\vec{\tilde{u}})}{k_F^{(t)}(\tilde{\vec{x}},\vec{\tilde{u}};\tilde{\vec{x}},\vec{\tilde{u}})} ( F^{(t)}(\tilde{\vec{x}},\vec{\tilde{u}}) - m_F^{(t)}(\tilde{\vec{x}},\vec{\tilde{u}}) )\bigg) \rho_\vec{U}(\vec{u}) d\vec{u},\\
&= m_Z^{(t)}(\xtarg) + \frac{ F^{(t)}(\tilde{\vec{x}},\vec{\tilde{u}}) - m_F^{(t)}(\tilde{\vec{x}},\tilde{\vec{u}})}{\sqrt{k_F^{(t)}(\tilde{\vec{x}},\vec{\tilde{u}};\tilde{\vec{x}},\vec{\tilde{u}})}} \frac{\int_{\mathbb{R}^m} k_F^{(t)}(\xtarg,\vec{u};\tilde{\vec{x}},\vec{\tilde{u}}) \rho_\vec{U}(\vec{u}) d\vec{u} }{\sqrt{k_F^{(t)}(\tilde{\vec{x}},\vec{\tilde{u}};\tilde{\vec{x}},\vec{\tilde{u}})}}, \\
&\sim \mathcal{N}\Bigg( m_Z^{(t)}(\xtarg) , \bigg(\frac{\int_{\mathbb{R}^m} k_F^{(t)}(\xtarg,\vec{u};\tilde{\vec{x}},\vec{\tilde{u}}) \rho_\vec{U}(\vec{u}) d\vec{u} }{\sqrt{k_F^{(t)}(\tilde{\vec{x}},\vec{\tilde{u}};\tilde{\vec{x}},\vec{\tilde{u}})}}\bigg)^2 \Bigg).
\end{split}
\end{equation*}

{
\section{Time complexity calculation details}
\label{app-timeCompl}
Again, recall that $t$ is the iteration number and the number of data points, $t\le \text{budget}$, $M$ is the number of $\mathbf u$ samples and $N$ is the number of GP trajectories in the simulations, $l$ the number of constraint GPs, $d$ and $m$ the dimensions of $\mathbf u$ and $\mathbf x$, respectively.
The steps for the derivation of the time complexity of the three algorithms are detailed below. It is assumed that the cost of the various internal optimizations is linear in the search space dimension.
\begin{itemize}
\item Learning of the $(l+1)$ GPs: $\mathcal{O}(t^3) \times (l + 1) \times (d+m)$, where the product by $(d+m)$, the input dimension size, accounts for the internal likelihood maximization.
\item Prediction of a GP (mean, variance, EI) at a point, assuming the covariance matrix was decomposed and inverted before: $\mathcal{O}(t^2)$ 
\item Sampling of a GP trajectory at $M$ points: $\mathcal{O}(M^3)$.
\item $\mathbb{P}(C^{(t)}(\mathbf{x}) \leq 0)$: it costs $N$ trajectories at $M$ points that is $N\times\mathcal{O}(M^3)$.
\item $\mean[ C^{(t)}(\vec{x}) ]$ for the calculation of $z^\text{feas}_\text{min}$: $M\times l$ predictions, i.e., $M \times l \times \mathcal{O}(t^2)$.
\item One calculation of EFI: the sum of the 2 aboves items plus the EI calculation, $N\times\mathcal{O}(M^3) + M \times l \times \mathcal{O}(t^2)$
\item The maximization of EFI: the input dimension times the above complexity but $z^\text{feas}_\text{min}$ is calculated only once, $d \times N\times\mathcal{O}(M^3) + M \times l \times \mathcal{O}(t^2)$, the dominant term is $d \times N\times\mathcal{O}(M^3)$.
\item Variance of the improvement: $N$ predictions, $N \times \mathcal{O}(t^2)$.
\item Constraints variance, $\int_{\mathbb{R}^m} p(\vec{u})(1-p(\vec{u}))\rho_\vec{U}(\vec{u}) d\vec{u}$: $M\times l$ predictions, $M\times l \times \mathcal{O}(t^2)$.
\item Sampling criterion: the sum of the 2 aboves items, $\max(N, M\times l) \times \mathcal{O}(t^2)$.
\item Minimization of the sampling criterion: the dimension of the uncertainties times the calculation of the sampling criterion, $m \times \max(N, M\times l) \times \mathcal{O}(t^2)$.
\end{itemize}
The overall time complexity of \EFISUR is $\max\left[\mathcal{O}(\text{budget}^3) \times (l + 1) \times (d+m) ~,~ d \times N\times\mathcal{O}(M^3)\right]$.
The complexity of \EFIRand is the same as that of \EFISUR because the GP learning and the EFI maximization which are common to both algorithms are driving it. The cost of the sampling criterion minimization is of second order.
Now, the cost of the \CEIDev algorithm is split as follows.
\begin{itemize}
\item Calculation of the quantiles : $M$ predictions of $l$ constraints, $M\times l \times \mathcal{O}(t^2)$.
\item Calculation of the EI : one calculation of $z^\text{feas}_\text{min}$ per iteration and a prediction of GP mean and variance.
\item Maximization of the quantile constrained EI: $d \times (M\times l + 2 ) \times \mathcal{O}(t^2)$.
\item Calculation of the deviation number, \DN: $l \times 2 \times \mathcal{O}(t^2)$.
\item Minimization of \DN: $m \times l \times 2 \times \mathcal{O}(t^2)$.
\end{itemize}
The complexity of \CEIDev is dominated by the cost of learning the GPs, $\mathcal{O}(\text{budget}^3) \times (l + 1) \times (d+m)$. If the number of uncertain parameters samples is larger than the budget of the GP regression, $M>\text{budget}$,
then \CEIDev will be faster to execute than \EFISUR and \EFIRand.
} % end rodoadd

\end{document}